\documentclass[10pt]{article}
\usepackage[letterpaper,textwidth=5.5in,textheight=9in,top=1in,headheight=12pt]{geometry}
\usepackage[T1]{fontenc}
\usepackage[utf8]{inputenc}
\usepackage{times}
\usepackage{microtype}
\usepackage{amsmath,amssymb}
\usepackage{booktabs,array,tabularx,multirow}
\usepackage{graphicx}
\usepackage{xcolor}
\usepackage[font=small,labelfont=bf,skip=4pt]{caption}
\usepackage{enumitem}
\setlist{nosep,leftmargin=*}
\usepackage[numbers,sort&compress]{natbib}
\usepackage[colorlinks=true,linkcolor=blue!55!black,citecolor=blue!55!black,urlcolor=blue!55!black]{hyperref}
\hypersetup{pdftitle={Generalization, Explainability and Limitation of Weight-Space Programming}}
\usepackage{titlesec}
\titleformat{\section}{\large\bfseries}{\thesection}{0.8em}{}
\titleformat{\subsection}{\normalsize\bfseries}{\thesubsection}{0.8em}{}
\titlespacing*{\section}{0pt}{14pt plus 2pt minus 2pt}{6pt}
\titlespacing*{\subsection}{0pt}{10pt plus 2pt minus 2pt}{4pt}
\usepackage{fancyhdr}
\newcommand{\web}{WEBench}
\newcommand{\astra}{Astra}
\newcommand{\E}{\mathbb{E}}
\newcommand{\R}{\mathbb{R}}
\newcommand{\relu}{\mathrm{relu}}

\providecommand{\Cov}{\operatorname{Cov}}
\providecommand{\Var}{\operatorname{Var}}
\providecommand{\He}{\mathrm{He}}
\newcommand{\casehead}[1]{\gdef\casetitle{#1}}
\newcommand{\casefacts}[6]{\par\medskip\noindent\begin{minipage}{\textwidth}\textbf{\casetitle}\par\smallskip
{\small\setlength{\tabcolsep}{4pt}\begin{tabularx}{\textwidth}{@{}>{\itshape}p{1.75cm}X@{}}\toprule
Task & #1\\ Network & #2\\ Given & #3\\ Hidden test & #4\\ Pass rule & #5\\ Answer set by & #6\\\bottomrule\end{tabularx}}\end{minipage}\par\smallskip}

\usepackage[normalem]{ulem}
\newif\ifreviewmode
\reviewmodefalse

\title{\bfseries Generalization, Explainability and Limitation of Weight-Space Programming}

\author{
Ali Sharifi-Zarchi \qquad Amir Bahmani\\[2pt]
{\small Stanford Deep Data Research Center}\\[-2pt]
{\small Department of Genetics, School of Medicine, Stanford University}\\[-2pt]
{\small\texttt{\{asharifi,abahman\}@stanford.edu}}
}

\date{}

\begin{document}
\maketitle
\thispagestyle{fancy}

\begin{abstract}
The surprising ability of large language models (LLMs) to work directly on neural network weights, known as weight-space programming, may reflect genuine generalization or familiarity with common architectures. Because prior evidence comes from a few experiments on standard designs, how far this capability extends to unfamiliar networks and tasks remains unclear. Here we introduce the Weight Engineering Benchmark (\web{}), covering 210 tasks across 56 families and 10 categories of operations for creating, modifying, and analyzing small networks, ensuring failures cannot be attributed to scale. Each family uses an iterative hardening ladder varying one design factor at a time, with at least three instances per task (mostly fresh networks). We evaluated GPT-6-Astra (Astra 6) once per instance with no training or fitting and at most 50 network runs. Astra 6 passed 1{,}218 of 1{,}461 scored instances (83\%), generalizing to rare sine and triangle-wave units, million-step sequences, and all 448 instances determined by weights alone. Many successes rely on human-verifiable mathematics, including the Chinese remainder theorem, covariances recovered from stored normalization statistics, and exact-arithmetic proofs. Yet generalization has a clear limit: Astra 6 failed in all 10 categories, and 8 of them contain tasks on which it failed the majority of the instances. Nearly all failures involve learned weights graded on unseen real data, and in the failures we analyzed, Astra 6 relied on unchecked false assumptions. On a 30-instance subset across ten tasks where Astra 6 performed worst, we evaluated seven additional models. Accuracy rose sharply from Astra 6 to Astra 6.2, reflecting rapid progress, with Claude Fable 5.2, Gemini 4 Neon, and ChatGPT Sol 6.1 trailing Astra 6.2 and all three open-weight models behind the commercial models. Every commercial model solved all constructed-weight instances, whereas unseen real-data tasks separated them sharply and open-weight models struggled even on constructed weights. Task statements, weights, and attempt records are available at \url{https://github.com/StanfordBioinformatics/WEBench}.
\end{abstract}

\section{Introduction}
\label{sec:introduction}

A neural network can be studied through the data used to train it or through the weights learned from those data.
Given an architecture and a loss function, an optimizer searches for weights that minimize the loss on the training data. This process can require substantial data, time, and computation.
A trained network may also need to be repaired, compressed, adapted, or merged after its original training data are no longer available. This makes it valuable to work directly with its weights, both to understand what it computes and to modify it without retraining.

Earlier work reaches weight space in two ways. \emph{Learned operators} are trained to produce, update or read the weights of other networks, including hypernetworks~\citep{fastweights,hypernet,t2l}, meta-learning~\citep{maml,l2l} and weight-space learning~\citep{weightspace,nfn,wslsurvey}. \emph{Task-specific methods} each solve one operation by a dedicated algorithm, including editing~\citep{rome}, merging~\citep{taskarith,rebasin}, compression~\citep{eckart,dfq,deepinv}, compilation of programs into weights~\citep{rasp,tracr}, verification~\citep{reluplex,crown} and interpretability~\citep{framework}. Learned operators require additional training, while task-specific methods focus on particular operations. This raises the question of whether a general-purpose model can work directly with network weights without being trained specifically for each task.

\citet{sparks} reported that a general-purpose language model could read, build and modify neural networks by working directly with their weights, without additional training for these tasks. This approach, which they called \emph{weight-space programming}, sparked a series of follow-up studies in robotics, quantum circuits, clinical networks and medical imaging~\citep{zhangrobot,vqc,clinical,medimg,eeg,hoo,iwsp,oksanen,gross}. Other groups scrutinized its successes~\citep{goworek,ahmed,kundu,liuyang,alug}, its failures~\citep{garrido,lehmann} and its benchmarking~\citep{iwibench,weightbench}. Of the two benchmarks, IWIBench~\citep{iwibench} tests edits of the behavior of a single language model, in 30 episodes from ten families, and WeightBench~\citep{weightbench} has 42 tasks in five tracks (reading, fixing, writing and compressing weights, and recognizing infeasible requests), each at one level of difficulty and 
with one or two independent agent attempts per task and model. Neither varies the difficulty of a task step by step, repeats a task on several networks, or sets aside tasks that a generic method solves.

How well these findings carry over to unfamiliar networks and tasks, and how the capability works, remains unclear. A model that rewrites the weights of an unseen network without data or retraining can easily appear opaque. We therefore ask three questions: (1) \emph{Generalization}: How far do a language model's weight-engineering abilities extend to architectures and tasks it is unlikely to have encountered? (2) \emph{Explainability}: To what extent can its techniques be understood and verified by a human engineer? (3) \emph{Limitations}: Where does a language model still fail even when network scale is not an obstacle?

We investigate these questions using \web{}, a benchmark of 210 tasks across 56 families covering ten weight operations. We evaluate GPT-6-Astra (\astra{}) across the full benchmark and compare seven additional language models on a 30-instance subset (Section~\ref{sec:models}). In each instance, working inside a fresh conversation, the solver receives a task statement and weight files, writes and runs analysis programs on those weights, and submits an answer. It has no external tools, may not train or fit parameters, and may run a network at most 50 times.

\textbf{Contributions.}
\begin{enumerate}
\item \emph{A benchmark,} \web{}, built with explicit qualification conditions so that both passes and failures can be interpreted cleanly.
\item \emph{Interpretable weight-engineering techniques used by \astra{},} explained and backed by proofs (Appendix~\ref{app:proofs}). To our knowledge, five have not previously been reported for a language model operating on weights: the Chinese remainder theorem for neural counters, input data statistics recovered from stored normalization parameters, multiscale error cancellation for counting, memory shared by two recurrent networks, and repair using network symmetry.
\item \emph{Interpretable patterns of failure,} including treating a non-linear network as linear, averaging weights where a maximum is required, judging repair sizes without measuring them, and mistaking similar weights for equivalent behavior. In five families with failures, programs of 2 to 30 lines pass most instances (and in a sixth, a procedure of about 40 lines solves many instances; Appendix~\ref{app:cases}).
\end{enumerate}

\section{Methods}
\label{sec:theory}
\subsection{General foundation}
\label{sec:spaces}
A neural network architecture $\mathcal A$ with $p$ parameters has the weight space $\Theta_{\mathcal A}=\R^{p}$. Each point $\theta\in\Theta_{\mathcal A}$ is one network, which computes a function $f_\theta:\mathcal X\to\mathcal Y$. Standard training reaches $\theta$ indirectly, by minimizing a loss on data. \emph{Weight engineering} works on the weights directly. A procedure $E$ creates a network from a specification, modifies given networks, or analyzes them to answer a question. Table~\ref{tab:cats} lists the ten operations covered by these three forms. Each form has a signature:
\begin{equation}
\begin{aligned}
\textbf{create:}\quad & E:\mathcal S\to\Theta_{\mathcal A} && \text{Construct}\\
\textbf{modify:}\quad & E:\Theta_{\mathcal A}^{k}\times\mathcal S\to\Theta_{\mathcal A'} && \text{Repair, Edit, Compress, Convert, Merge, Adapt}\\
\textbf{analyze:}\quad & E:\Theta_{\mathcal A}^{k}\times\mathcal S\to\mathcal O && \text{Read, Diagnose, Verify}
\end{aligned}
\label{eq:ops}
\end{equation}
Here $\mathcal S$ is the set of specifications. A specification states what is asked, for example a function to compute, a size limit or a question, and it may prescribe parts of the network. The set $\mathcal O$ holds answers about networks, such as a property, a diagnosis or a certificate. In the second and third forms, $k\ge 1$ networks are supplied. They may have different architectures $\mathcal A_1,\dots,\mathcal A_k$, in which case $\Theta_{\mathcal A}^{k}$ stands for the product $\Theta_{\mathcal A_1}\times\dots\times\Theta_{\mathcal A_k}$. The architecture $\mathcal A'$ of the result may equal $\mathcal A$, as in Repair and Edit, or differ from it, as in Compress and Convert.

\begin{table}[!tb]
\centering\footnotesize
\caption{The ten operations of \web{}, each with a representative formal statement and an example, in the order of the three forms of Eq.~\eqref{eq:ops}. Supplied weights are $\theta$ (or $\tilde\theta$ when they were damaged from $\theta^\star$), submitted weights are $\theta'$, 
and $f\approx g$ denotes the agreement required by the task, which may involve numerical output error, matching predictions, or a performance threshold on test data.
}
\label{tab:cats}
\setlength{\tabcolsep}{4pt}
\begin{tabularx}{\textwidth}{@{}p{1.2cm}Xp{4.7cm}@{}}
\toprule
Operation & Representative formal statement & Example \\
\midrule
Construct & Given a specification $s\in\mathcal S$ that states a target function $g$ and may fix parts of the network, output $\theta'\in\Theta_{\mathcal A}$ with $f_{\theta'}\approx g$. & Compile a program over binary signals into a recurrent ReLU network that follows it for $10^{5}$ steps. \\
Repair & The supplied weights $\tilde\theta=C(\theta^\star)$ were damaged by an unknown corruption $C$ of a known kind; output $\theta'$ with $f_{\theta'}\approx f_{\theta^\star}$. & Undo further training of a classifier's last layers on wrongly labeled photographs. \\
Edit & Given a set $U\subset\mathcal X$ of inputs and a target behavior $h$, output $\theta'$ with $f_{\theta'}\approx h$ on $U$ and $f_{\theta'}\approx f_\theta$ on $\mathcal X\setminus U$, under a limit on which weights may change or on the rank of the change. & Make a 20-breed classifier assign photographs of one breed to another. \\
Compress & Output $\theta'\in\Theta_{\mathcal A'}$ with $f_{\theta'}\approx f_\theta$ for a smaller architecture $\mathcal A'$, given by a cap on rank, width, depth or bits. & Replace the input layer of a VAE encoder by a layer of rank $r$. \\
Convert & Output $\theta'\in\Theta_{\mathcal A'}$ that reproduces the source network's behavior in a specified target representation, allowing for changes in architecture, activation functions, or input and output coordinates. & Turn a hand-built recurrent ReLU network into a GRU with a stated number of units. \\
Merge & Given $\theta_1,\dots,\theta_k$ with $k\ge2$, output the weights $\theta'$ of one network of a stated size that reproduces every source: $f_{\theta'}(x)\approx\big(f_{\theta_1}(x),\dots,f_{\theta_k}(x)\big)$, or $f_{\theta'}(x,j)\approx f_{\theta_j}(x)$ when an input $j$ selects the source, in each case on the domain of the source concerned. & One GRU that reproduces the outputs of two 64-unit GRUs at every step. \\
Adapt & The conditions of deployment change, typically the inputs by a map $\tau$ that may be specified or estimated from unlabeled examples; output $\theta'$ with $f_{\theta'}(\tau(x))\approx f_\theta(x)$. & Keep a photo classifier's predictions when the camera response changes. \\
Read & Output a task-defined property
 $a=R(\theta)\in\mathcal O$ of the network, with a witness where one exists, such as $\max_{x\in\mathcal X}f_\theta(x)$ and a point that attains it. & Maximum of a six-layer ReLU network over $[-1,1]^{12}$, with a point that attains it. \\
Diagnose & The supplied weights $\tilde\theta=C_\xi(\theta^\star)$ were damaged by a fault $\xi$; output $\xi$, that is, its location, type or magnitude. & Identify which label groups were merged when a photo classifier was fine-tuned. \\
Verify & Decide whether a stated claim about one or more networks holds. Where the task asks for it, a false claim is refuted by a counterexample, such as a point $x$ with $\|f_{\theta_1}(x)-f_{\theta_2}(x)\|>\varepsilon$, and a true claim is supported by a certificate that the grader can check, such as an upper bound on $\max_{x\in\mathcal X}\|f_{\theta_1}(x)-f_{\theta_2}(x)\|$. & Decide whether two ReLU networks agree on $[-1,1]^{16}$, with a checkable certificate of agreement or a point of disagreement. \\
\bottomrule
\end{tabularx}
\end{table}

\subsection{Evaluation framework}
\label{sec:categories}
\label{sec:def}
\textbf{Definition 1 (task).} A task has a generator that produces instances $(\mathcal I,\xi)$. The information $\mathcal I=(s,\theta_1,\dots,\theta_k)$ is given to the solver, including a specification $s$, which includes any examples or statistics that are supplied, and the weights of $k\ge 0$ networks. The material $\xi$ is hidden and used for grading, for example held-out inputs or the location of a planted fault. The solver $E$ returns an answer $o=E(\mathcal I)$ from a set $\Omega$ of admissible answers, which is a weight space for the first two forms of Eq.~\eqref{eq:ops} and the set $\mathcal O$ for the third. A grader $\Phi(o;\mathcal I,\xi)\in\{0,1\}$ accepts the answer if it meets every pass threshold of the task; we call these thresholds the \emph{pass rule}. The solver is restricted to a set $\mathcal M$ of permitted procedures. A procedure in $\mathcal M$ may compute on the supplied weights and may observe the outputs of a network in at most 50 \emph{runs} on inputs of its choice. A run is one request to execute a network, and computing the output or a hidden activation of a network for a specific input in any other way, in the solver's own program or by hand, is not permitted. It may not \emph{fit}, that is, choose its answer by reducing an error measured on the outputs of a network, as training, regression on outputs and distillation do. Choosing a discrete answer from the outputs observed in the permitted runs, for example which output stands for which breed, is not fitting.

\noindent\textbf{Definition 2 (qualified instance).}
 An instance $(\mathcal I,\xi)$ of a task is qualified if the following hold.
\begin{enumerate}
\item[(a)] \emph{Reference.} A stored answer $o^{\mathrm{ref}}\in\Omega$ satisfies $\Phi(o^{\mathrm{ref}};\mathcal I,\xi)=1$; it may be produced by any means, including training on data that the solver lacks, and therefore shows only that a passing answer exists.
\item[(b)] \emph{No bypass.} Neither $\xi$ nor $o^{\mathrm{ref}}$ leaks through a route other than the intended one, such as extraneous files, metadata, the way in which values are stored, or a defect of the grader. In addition, an uninformed guessing strategy, which is specified for each task, passes with probability below 1\%.
\item[(c)] \emph{Baseline survival.} A fixed suite $\mathfrak B\subset\mathcal M$ of simple procedures fails, that is, $\Phi(b(\mathcal I);\mathcal I,\xi)=0$ for every $b\in\mathfrak B$. The suite covers trivial answers, standard weight statistics and decompositions, generic symbolic analysis, probing within the permitted runs, and generic sweeps over an edit, and is fixed in advance on separate development instances.
\item[(d)] \emph{Feasibility.} A written argument explains how some procedure $E^\star\in\mathcal M$ reaches a passing answer from $\mathcal I$ and the observations permitted by $\mathcal M$, naming the task-specific step required.
\end{enumerate}
On a qualified instance, conditions (b) and (c) rule out leaks, guessing, and simple baselines as explanations for a pass, while conditions (a) and (d) justify reading a failure as a limit of the solver rather than a flaw in the task. We \emph{score} an instance as follows. A pass is scored if the attempt followed the rules and the instance meets conditions (a) through (c). A miss is scored as a failure only if, in addition, a solution using permitted procedures has been demonstrated to pass on that instance; the written argument of condition (d) alone is not enough (Section~\ref{sec:limits}).

\subsection{Workflow}
\label{sec:workflow}
\label{sec:setup}
We developed a workflow that turns these definitions into a benchmark in four steps. Three terms are used throughout. A \emph{family} is a set of related tasks that pose one kind of problem. Its tasks are versions of that problem that differ in difficulty, usually in a single design factor such as a size limit, the severity of damage, or the amount of evidence supplied. An \emph{instance} is one concrete realization of a task, drawn by the task's generator, with its own network or networks, the information given to the solver, and its own hidden test. Every task has at least three scored instances, in most tasks different networks, so that a result does not rest on a single network. Each instance is attempted once, in a fresh conversation with no memory of the other attempts, so that results can be counted in instances. In an attempt the solver may make at most 50 \emph{runs}, each an execution of a network on inputs of its choice.

\textbf{Step 1: build and qualify.} 
For each family, the workflow uses Claude Fable 5.2 to help produce the task statement, the generator and the grader, and then establishes the four conditions of Definition~2, namely a reference answer (a), an audit for leaks and a guessing check (b), the fixed baselines (c), and a solution by permitted procedures (d).

\textbf{Step 2: harden.} A \emph{ladder} is a sequence of tasks $T_{\lambda_1},\dots,T_{\lambda_m}$ of one family that differ in a single design factor $\lambda$, stated in advance, with instances matched across levels. The \emph{success curve} of a solver $E$ restricted to $\mathcal M$ is
\begin{equation}
p_E(\lambda)=\Pr\big[\Phi(E(\mathcal I);\mathcal I,\xi)=1\big],
\label{eq:curve}
\end{equation}
estimated by the fraction of passes among the scored instances of a level. 
We vary the factor across qualified levels to find where the solver's pass rate declines. In some families it declines gradually, and in others it does not decline within the range that we built; we do not claim a sharp boundary. The tasks are thus selected adaptively with respect to the evaluated model~\citep{levitt,dynabench,perez}.

\textbf{Step 3: evaluate and audit.} Every instance is attempted once with our harness, and each is also released as a Terminal-Bench task~\citep{tbench}. Of 433 tasks created, 281 in 81 families were evaluated on 2{,}118 instances and audited again, with a second language model reviewing every record. \web{} keeps the 210 tasks in 56 families that have at least three scored instances; 1{,}461 instances are scored in all.

\textbf{Protocol and data.} The evaluated model is \texttt{gpt-6-astra}; Section~\ref{sec:models} compares seven further language models with it on 30 instances of ten tasks. In an attempt it works in a sandbox with Python and numerical libraries, two CPU threads, 8\,GB of memory, no GPU and no internet. It may send at most 90 replies, each with one action, namely a program of its own, a run of a network, a format check of its answer file or the submission, within 60 minutes of tool time, and it receives no feedback from the hidden test. An attempt used a median of 9 replies, and the 2{,}118 attempts took 659 hours in all. The learned networks were trained on photographs of the Oxford-IIIT Pet dataset~\citep{parkhi}, reduced to $64\times64$ pixels, on 
the 1{,}797-image, $8\times8$ scikit-learn digits subset of the UCI dataset~\citep{optdigits} or on simulated data, and every hidden test uses data that were set aside before training.

\textbf{Step 4: analyze.} We group the families by what determines the correct answer, coded from the task descriptions. In 26 families it is \emph{the weights alone}: the answer can be computed, and checked, exactly from the supplied weights and the specification. In 7 families it is \emph{the weights with a stated statistical model}: the task statement supplies the statistical assumptions or the generating procedure that, together with the weights, determine the answer, as when the inputs are stated to be Gaussian (Case~S6) or a planted fault of a stated kind must be named, and no hidden real data enter the grading. In 23 families it is \emph{behavior on unseen real data}: the answer is graded on real data that the solver does not receive, such as held-out photographs. We analyze the programs that \astra{} wrote for every instance and support the case studies by \emph{controlled comparisons} between levels of a ladder, \emph{one-change reruns} of a submitted program and \emph{frozen-program reruns} on other instances.

\section{Results}
\label{sec:results}

\subsection{Overview}
\label{sec:map}

\begin{table}[!tb]
\centering\footnotesize
\caption{Results of \astra{} on the 210 tasks of \web{}, by operation. \emph{Instances}: scored instances. \emph{Passed}: scored instances whose submitted answer passed the hidden test. The three columns under \emph{Answer determined by} split the scored instances by what fixes the correct answer (Section~\ref{sec:workflow}): the supplied weights alone; the weights together with a statistical model stated in the task; or the behavior of the network on real data that \astra{} does not receive. Each of these entries is passed/scored. \emph{Failed}: scored instances that did not pass. \emph{Families with a failure}: families of the operation with at least one failed instance, out of all families of the operation.}
\label{tab:ops}
\setlength{\tabcolsep}{2.7pt}
\begin{tabular}{@{}lrrrrrrrrr@{}}
\toprule
 & & & & & \multicolumn{3}{c}{Answer determined by} & & Families\\
\cmidrule(lr){6-8}
Operation & Families & Tasks & Instances & Passed & weights alone & weights + model & unseen data & Failed & with a failure\\
\midrule
Construct & 7 & 23 & 126 & 104 (83\%) & 85/85 & -- & 19/41 & 22 & 2/7 \\
Repair & 4 & 22 & 138 & 84 (61\%) & 59/59 & -- & 25/79 & 54 & 2/4 \\
Edit & 4 & 10 & 58 & 38 (66\%) & 15/15 & -- & 23/43 & 20 & 3/4 \\
Compress & 4 & 17 & 128 & 115 (90\%) & 20/20 & 20/20 & 75/88 & 13 & 1/4 \\
Convert & 13 & 35 & 236 & 226 (96\%) & 130/130 & -- & 96/106 & 10 & 2/13 \\
Merge & 6 & 29 & 233 & 215 (92\%) & 55/55 & 70/70 & 90/108 & 18 & 1/6 \\
Adapt & 5 & 20 & 163 & 156 (96\%) & 20/20 & 47/49 & 89/94 & 7 & 3/5 \\
Read & 6 & 28 & 186 & 125 (67\%) & 40/40 & 20/20 & 65/126 & 61 & 3/6 \\
Diagnose & 4 & 16 & 121 & 108 (89\%) & 9/9 & 99/112 & -- & 13 & 3/4 \\
Verify & 3 & 10 & 72 & 47 (65\%) & 15/15 & -- & 32/57 & 25 & 1/3 \\
\midrule
\textbf{All} & \textbf{56} & \textbf{210} & \textbf{1{,}461} & \textbf{1{,}218 (83\%)} & \textbf{448/448} & \textbf{256/271} & \textbf{514/742} & \textbf{243} & \textbf{21/56} \\
\bottomrule
\end{tabular}
\end{table}

\astra{} passed 1{,}218 of the 1{,}461 scored instances, that is 83\% (Table~\ref{tab:ops}). Because the instances of one family come from the same generator, they are not independent, so we estimated the uncertainty by resampling whole families rather than single instances. This gives a 95\% interval of 75\% to 91\% for the overall pass rate. On the retained tasks, \astra{} also missed 149 instances that we do not score, because no solution using permitted procedures has been demonstrated for them. Counting these misses as failures would lower the pass rate to 76\% (Section~\ref{sec:limits}).

The failures are concentrated in a few families. Of the 56 families, 35 have no failure at all and 21 have at least one, and six of them hold 146 of the 243 failures. Sections~\ref{sec:general} to~\ref{sec:contrasts} take up the three questions of the Introduction, and Section~\ref{sec:models} compares \astra{} with seven other language models:
\begin{itemize}
\item \textbf{Generalization (Section~\ref{sec:general}).} \astra{} passed all 448 instances in which the supplied weights alone determine the answer, including networks with rarely used activation functions and computations that run for a million steps.
\item \textbf{Explainability (Section~\ref{sec:success}).} Many of its successful solutions use techniques that a reader can follow step by step and check.
\item \textbf{Limitations (Section~\ref{sec:contrasts}).} Almost all failures (241 of 243) occur on networks whose weights were learned by training on data, and most of them (228) also depend on how the network behaves on real data that \astra{} does not see. On 30 tasks, \astra{} failed more than half of the scored instances.
\item \textbf{Other language models (Section~\ref{sec:models}).} On 30 instances of ten tasks, every commercial model passes all 9 constructed-weight instances while accuracy on unseen photographs ranges from 1/15 (\astra{}) to 14/15 (\astra{} 6.2; 29/30 overall), and the three open-weight models pass at most 6.
\end{itemize}

\subsection{Generalization}
\label{sec:general}

\textbf{Every instance passed when the weights alone determine the answer.} In 448 scored instances, spread over all ten operations, the correct answer can be computed exactly from the supplied weights, and \astra{} passed all 448. It also passed all 478 instances with \emph{constructed weights}, which we wrote with a program instead of training. Such networks can still be hard to read. One of our generators, for example, writes a recurrent network with several counters and a small state machine and then shuffles and rescales its units, so that the design cannot be read off the weights.

\textbf{Rare activation functions.} Success on common architectures could rest on familiarity with them. To check this, we built three problems in two versions each, one with common units and one with sine or triangle-wave units, which are rarely used in practice. In the first problem, an arithmetic program is compiled into a recurrent network. In the second, a graph network is rewritten so that it still works when the nodes of its input are reordered and its features are rotated. In the third, a convolutional network is rewritten for a signal whose even and odd time steps are stored in separate channels. \astra{} passed all 15 instances with the rare units.

\textbf{Networks rebuilt from unfamiliar layers (Case~S10).} In this task, the layers of a network are fixed in advance and contain no ReLU. Instead, they use the absolute value, a triangle wave, the sine or products of two units. \astra{} had to choose the weights of this network so that it gives the same outputs as a given ReLU network. It wrote each ReLU as a combination of absolute values, and it sent signals through the smooth layers at such a small scale that these layers behave like straight lines (Results~\ref{r:abs} and~\ref{r:wire}). All 20 instances passed. \astra{} also passed all 62 scored instances of two related families on learned weights, one with absolute-value units and one with Snake units.

\textbf{Very long sequences (Case~S4).} \astra{} rewrote recurrent counting networks as GRUs that reproduce the original outputs at every step of sequences of up to about one million steps, although one permitted run of the network covers at most 4{,}096 steps. Across the instances, it found five different designs that keep the counts exact over the whole length. In three instances, it stored the same count at several small scales and combined them so that the errors caused by the bending of the saturating units cancel out, which is a form of Richardson extrapolation (Results~\ref{r:lanes} to~\ref{r:exact}). All 40 instances passed.

\textbf{Learned weights with no data or only a few unlabeled examples.} \astra{} also succeeded on several tasks with learned weights in which it received no data, or only a few examples without labels. Without seeing a single photograph, it adapted a 12-breed classifier to photographs in which 3 to 7 of the 12 color values of every $2\times2$ block are deleted, and it passed 49 of 50 instances (Case~S1). Also without data, it pruned the last convolutional layer of a classifier from 128 channels to as few as 12 and passed all 50 instances (Case~S2). In a third task, a new sensor rotates, scales and shifts its readings by unknown amounts, and \astra{} had to adapt a classifier of sensor readings to this sensor. It received at most 64 unlabeled readings, and in some versions as few as 4 readings or a second network trained on another task, and it passed 47 of 49 instances (Case~S12). Section~\ref{sec:success} explains the mathematics behind these solutions.

\begin{figure}[!tb]
\centering
\includegraphics[width=0.6\textwidth]{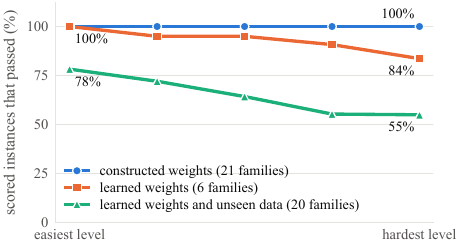}
\caption{Pass rate along the difficulty ladders of the 47 families with more than one level (1{,}353 scored instances), from the easiest to the hardest level of the factor that each family varies. Each line is the mean over the families of one group; groups, colors and markers are those of Figure~\ref{fig:map}. Versions of a family that vary a different factor are left out.}
\label{fig:ladders}
\end{figure}

\textbf{Along the difficulty ladders, constructed weights stay at 100\% and learned weights decline.} Most families form a ladder of levels that make one design factor harder step by step. Figure~\ref{fig:ladders} follows the 47 families with more than one level from their easiest to their hardest level. On constructed weights, the pass rate stays at 100\% at every level (21 families). On learned weights where the answer does not depend on unseen data, the average pass rate falls from 100\% to 84\% (6 families). Where the answer does depend on unseen data, it falls from 78\% to 55\% (20 families), and the hardest level is lower than the easiest in 13 of these families, equal in 6 and higher in 1. Because we built the levels to find where the pass rate declines (Section~\ref{sec:workflow}), the figure shows where the decline occurs, but it is not a statistical test of the decline.

\subsection{Explainability}
\label{sec:success}

To learn how \astra{} succeeds, we had two other models, Claude Fable 5.2 and ChatGPT Sol 6.1 (\texttt{gpt-6.1-sol}), annotate the complete records of 2{,}088 attempted instances. Each record was annotated at three levels, namely the single instance, the task family and the operation (Appendix~\ref{app:cases}).

\textbf{Many passes needed no run of the network.} On 570 of the 1{,}218 passed instances, almost half, \astra{} did not run the network at all. It worked from the weights, the task statement and mathematics alone, and on some of these instances it proved that its answer is correct instead of testing it.

\textbf{Fourteen groups of techniques.} We sorted the methods that \astra{}'s programs use into 14 groups (Table~\ref{tab:matrix}). Many of these techniques can be explained in the sense that a reader can follow every step and check it, as in the following three examples.

\textbf{Modular arithmetic instead of simulation (Case~S3).} A recurrent network of about 20 units contains six counters, and \astra{} must give the exact number of steps after which each of the network's four outputs repeats. This number can exceed $10^{13}$, whereas one permitted run of the network shows only 1{,}024 steps. \astra{} read each counter and its period directly from the weights, wrote the conditions under which the outputs repeat as a system of congruences, and solved it with the Chinese remainder theorem (Result~\ref{r:clock}). It passed all 20 instances without running the network once.

\textbf{Showing that ReLU units never cut anything off (Case~S8).} Two photo classifiers share one image encoder, and each has a head of 128 hidden ReLU units. Without seeing any photograph, \astra{} must merge the two heads into one much smaller head that agrees with both classifiers on hidden photographs. It noticed that the features coming out of the encoder are never negative, and that about 110 units of each head have no negative weight and no negative bias (the task statement says that some units were trained with non-negative weights). The input to such a unit is therefore never negative, so its ReLU never cuts anything off and the unit is linear on every possible photograph. Together, these units collapse into one small linear map (Result~\ref{r:on}). Because this holds for every photograph, the merged head needed no test on data, and all 70 instances passed.

\textbf{Reading image statistics from a normalization layer (Case~S1).} In this task, 3 to 7 of the 12 color values in every $2\times2$ block of a photograph are deleted, and \astra{} must adapt a 12-breed classifier to such photographs without seeing any of them. It noticed that the first normalization layer stores, for each of the 96 first-layer filters, the variance of the filter's output over the training photographs. Each of these variances is a linear equation in the covariance of the color values that the filters read, so \astra{} solved the 96 equations for that covariance. From the covariance, it computed the best linear estimate of the deleted color values from the remaining ones and wrote this estimate into the weights of the first layer (Results~\ref{r:var} and~\ref{r:lmmse}). It passed 49 of 50 instances.

\textbf{Further techniques.} Other successful techniques can be explained in the same way, among them a Gaussian model of the units for pruning without data (Case~S2), error cancellation across scales in a GRU (Case~S4), a memory shared by two recurrent networks when they are merged (Case~S5), repair by the symmetry of a network (Case~S7) and proofs in exact fractions (Case~S9). Appendix~\ref{app:cases} describes each case, Appendix~\ref{app:proofs} gives the mathematics, and Table~\ref{tab:tech} lists the most striking techniques with the number of instances on which they were used.

\begin{table}[!tb]
\centering\footnotesize
\caption{Groups of techniques, by operation, from the annotation of 2{,}088 instances, including instances that are not scored. Each number sums, over the techniques of the group, the instances on which each technique was used, so an instance on which several techniques were used is counted more than once. The symbol gives the outcome according to the annotation: $\bullet$, every technique of the group led to passes in that operation; $\circ$, none of them led to a pass; $\odot$, mixed. An empty cell means that the group was not used in that operation. Appendix~\ref{app:cases} explains how the groups were formed.}
\label{tab:matrix}
\setlength{\tabcolsep}{2.6pt}
\begin{tabularx}{\textwidth}{@{}X*{10}{c}@{}}
\toprule
Group of techniques & \rotatebox{60}{Construct} & \rotatebox{60}{Repair} & \rotatebox{60}{Edit} & \rotatebox{60}{Compress} & \rotatebox{60}{Convert} & \rotatebox{60}{Merge} & \rotatebox{60}{Adapt} & \rotatebox{60}{Read} & \rotatebox{60}{Diagnose} & \rotatebox{60}{Verify}\\
\midrule
Exact arithmetic and proofs & $\odot$\,{\scriptsize 56} & $\odot$\,{\scriptsize 8} & $\odot$\,{\scriptsize 15} & $\bullet$\,{\scriptsize 22} & $\bullet$\,{\scriptsize 6} &  & $\odot$\,{\scriptsize 23} & $\bullet$\,{\scriptsize 66} & $\odot$\,{\scriptsize 33} & $\odot$\,{\scriptsize 32} \\
Modular arithmetic & $\bullet$\,{\scriptsize 18} &  &  & $\bullet$\,{\scriptsize 49} & $\bullet$\,{\scriptsize 64} &  &  & $\bullet$\,{\scriptsize 20} &  &  \\
Logic gates built from units & $\odot$\,{\scriptsize 55} & $\bullet$\,{\scriptsize 25} & $\odot$\,{\scriptsize 76} & $\circ$\,{\scriptsize 1} & $\odot$\,{\scriptsize 36} &  & $\odot$\,{\scriptsize 3} &  &  &  \\
Identities between activation functions & $\odot$\,{\scriptsize 14} & $\circ$\,{\scriptsize 2} & $\odot$\,{\scriptsize 11} & $\odot$\,{\scriptsize 6} & $\odot$\,{\scriptsize 194} & $\odot$\,{\scriptsize 127} & $\circ$\,{\scriptsize 1} &  &  &  \\
Recovery of hidden structure &  &  & $\odot$\,{\scriptsize 3} &  & $\bullet$\,{\scriptsize 6} & $\bullet$\,{\scriptsize 96} & $\bullet$\,{\scriptsize 20} &  &  & $\bullet$\,{\scriptsize 80} \\
Stored data statistics & $\circ$\,{\scriptsize 17} &  &  & $\odot$\,{\scriptsize 72} & $\odot$\,{\scriptsize 66} & $\odot$\,{\scriptsize 11} & $\odot$\,{\scriptsize 117} & $\odot$\,{\scriptsize 21} & $\odot$\,{\scriptsize 36} &  \\
Gaussian models of units & $\circ$\,{\scriptsize 7} &  & $\odot$\,{\scriptsize 4} & $\odot$\,{\scriptsize 138} & $\odot$\,{\scriptsize 10} & $\bullet$\,{\scriptsize 6} & $\odot$\,{\scriptsize 9} &  &  & $\circ$\,{\scriptsize 12} \\
Closed-form inversion &  &  &  &  &  &  & $\odot$\,{\scriptsize 56} &  &  &  \\
Probability models and sampling &  & $\odot$\,{\scriptsize 57} & $\odot$\,{\scriptsize 26} & $\bullet$\,{\scriptsize 20} &  &  & $\odot$\,{\scriptsize 109} &  & $\odot$\,{\scriptsize 22} &  \\
Probe inputs &  &  & $\odot$\,{\scriptsize 3} & $\bullet$\,{\scriptsize 24} & $\odot$\,{\scriptsize 10} & $\circ$\,{\scriptsize 3} &  & $\odot$\,{\scriptsize 85} & $\odot$\,{\scriptsize 41} &  \\
Exhaustive enumeration &  &  &  &  &  &  &  & $\odot$\,{\scriptsize 34} & $\odot$\,{\scriptsize 33} &  \\
Weight similarity & $\odot$\,{\scriptsize 23} & $\odot$\,{\scriptsize 110} & $\odot$\,{\scriptsize 43} & $\circ$\,{\scriptsize 10} & $\odot$\,{\scriptsize 43} & $\odot$\,{\scriptsize 30} & $\odot$\,{\scriptsize 1} & $\odot$\,{\scriptsize 148} & $\odot$\,{\scriptsize 50} & $\circ$\,{\scriptsize 34} \\
Prior knowledge & $\odot$\,{\scriptsize 11} &  & $\circ$\,{\scriptsize 3} & $\circ$\,{\scriptsize 14} &  &  &  & $\odot$\,{\scriptsize 9} &  &  \\
Leaks in the files &  & $\circ$\,{\scriptsize 11} &  &  &  & $\circ$\,{\scriptsize 2} &  & $\odot$\,{\scriptsize 20} & $\odot$\,{\scriptsize 21} & $\odot$\,{\scriptsize 197} \\
\bottomrule
\end{tabularx}
\end{table}

\subsection{Weight engineering limitations}
\label{sec:contrasts}
\label{sec:synthesis}
\label{sec:mechanisms}

\begin{figure}[!tb]
\centering
\includegraphics[width=0.82\textwidth]{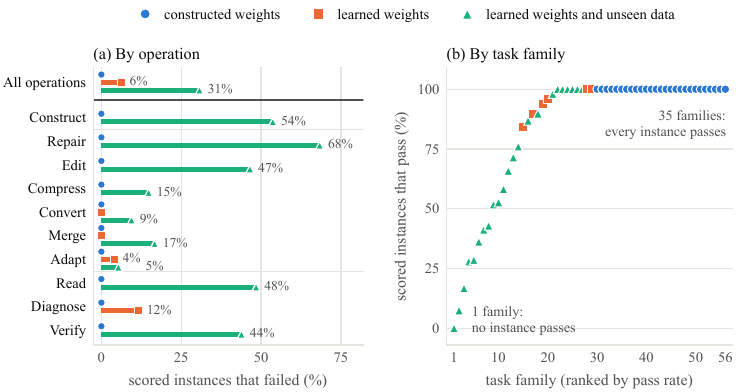}
\caption{Results of \astra{} by the two features that almost all failures share: learned weights, and a correct answer that depends on unseen data. Colors and markers denote the same three groups in both panels. (a) Share of scored instances that failed in each group, over all operations and within each operation (operations ordered as in Eq.~\eqref{eq:ops}). A group with no failure appears as a marker at 0\%; a group is absent where the operation has no tasks of that group. (b) Pass rate of each of the 56 families, from lowest to highest, one marker per family. The one family in which no weights are supplied and \astra{} writes a photo classifier from the class names alone (7 instances, 2 failures) is counted with the group whose answer depends on unseen data.}
\label{fig:map}
\end{figure}

\textbf{Two features are shared by almost all failures.} Almost all of the 243 failures have the following two features.
\begin{enumerate}
\item \textbf{Learned weights.} 241 of the 243 failures are on networks whose weights were learned by training on data. The remaining two occur in the single family where no weights are supplied and \astra{} constructs a photo classifier solely from the class names. No instance with constructed weights failed, in any operation (Figure~\ref{fig:map}a).
\item \textbf{Unseen data.} In 228 of the 243 failures (94\%), the correct answer depends on how the network behaves on real data that \astra{} does not see, such as the hidden test photographs.
\end{enumerate}
The failure rate is highest when both features are present. Where the weights were learned but the answer does not depend on unseen data, \astra{} failed 6\% of the instances, with a 95\% interval of 1\% to 12\% from resampling families. Where both features are present, it failed 31\% of the instances, with an interval of 18\% to 45\%. Even so, \astra{} passed 69\% of the instances that have both features, so neither feature makes a task impossible.

\textbf{The pattern does not depend on how borderline families are assigned.} Six families could reasonably be put into a different one of the three groups of Table~\ref{tab:ops} (Appendix~\ref{app:cases}). If all six are moved, the three groups have 580 of 580, 157 of 159 and 481 of 722 instances passed, and 241 of the 243 failures fall in the group whose answer depends on unseen data.

\textbf{Many failures occurred without a network run.} In 163 of its 243 failures, \astra{} did not run the network at all. It also made no run in 570 of its 1{,}218 passes, so skipping the runs is not a mistake in itself, because a run adds nothing when exact mathematics settles the answer. Where the answer depends on unseen data, however, the difference is large. There, 153 of the 370 instances without a run failed (41\%), compared with 75 of the 372 instances with at least one run (20\%). Since these are different tasks and not a controlled experiment, the comparison shows an association and not the effect of running the network (Appendix~\ref{app:cases}).

\textbf{On 30 tasks, \astra{} failed most instances.} On 30 of the 210 tasks, in 13 families and 8 of the ten operations, \astra{} failed more than half of the scored instances, and these tasks hold 168 of the 243 failures (Table~\ref{tab:majority}). In eight of the 13 families, another version of the same task was mostly passed. The two versions can differ only slightly, for example by a layer of rank 32 instead of 24, a pass mark of 65\% instead of 70\%, or half a round of faulty training instead of one.

\begin{table}[!tb]
\centering\footnotesize
\caption{The 13 task families that contain tasks on which \astra{} failed more than half of the scored instances: 30 tasks in 8 operations, holding 168 of the 243 failures (Adapt and Diagnose have none). \emph{Mostly failed at}: the versions of the family on which this happened. \emph{Tasks}: their number. \emph{Passed}: passed/scored instances over these tasks. \emph{Case}: the case study in Appendix~\ref{app:cases}. No photographs are supplied unless the family says so, and every answer is graded on real images that \astra{} does not receive.}
\label{tab:majority}
\setlength{\tabcolsep}{4pt}
\begin{tabularx}{\textwidth}{@{}p{1.15cm}X>{\raggedright\arraybackslash}p{3.1cm}rrl@{}}
\toprule
Operation & Task family & Mostly failed at & Tasks & Passed & Case \\
\midrule
Construct & Give a 20-breed classifier new classes that join several breeds, by rewriting its final layer. & 4 or 10 new classes & 2 & 2/17 & F1 \\
 & Write a small classifier of two named kinds of photographs from the names alone. & a pass mark of 70\% & 1 & 1/3 & -- \\
Repair & Undo further training of a classifier's last layers on wrongly named photographs, with two groups of look-alike breeds. & 1, 2, 4 or 8 rounds of such training & 4 & 8/27 & F4 \\
 & The same repair with three groups of look-alike breeds. & 2, 4, 8 or 12 rounds, and 12 rounds on an earlier set of networks & 5 & 12/43 & F4 \\
Edit & Make a cat-or-dog classifier call pugs cats by a low-rank change of one layer, with all other predictions unchanged. & its single task & 1 & 0/4 & -- \\
 & Redirect breeds to their look-alike partners by rewriting the final layer of a 20-breed classifier. & 5 or 10 breeds & 2 & 4/14 & F1 \\
Compress & Lower the rank of the first layer of a digit encoder, given 64 example images. & rank 24 & 1 & 1/8 & F3 \\
Convert & Replace the first-layer filters of a photo classifier by filters of a single row. & its single task & 1 & 1/6 & F8 \\
Merge & Merge two specialist classifiers into one network of the same size, when their idle units were trained to mislead. & the two strongest levels of that training & 2 & 4/19 & F2 \\
Read & Find the output that a classifier favors for each breed after its training labels were partly mixed, given a few labeled photographs. & 8, 4, 2 or 1 photographs per breed & 4 & 11/33 & F7 \\
 & The same without photographs. & the earliest set of networks & 1 & 2/5 & F7 \\
 & Name the shuffled outputs of a 12-breed classifier. & 3 or no names given & 2 & 2/12 & F6 \\
Verify & Decide which of ten proposed repairs restore the accuracy of a damaged classifier. & proposals 6, 5, 4 or 3 points from the threshold & 4 & 2/27 & F5 \\
\midrule
\textbf{All} & & & \textbf{30} & \textbf{50/218} & \\
\bottomrule
\end{tabularx}
\end{table}

\textbf{Why \astra{} failed: unchecked assumptions.} The failures we examined in detail are just as interpretable as the successes. In each case, \astra{} committed to a flawed assumption or heuristic and either skipped checking its answer or ran a check that could not detect the error. We review two cases below as concrete examples; six further failure families (Cases~F3 to~F8) are analyzed in Appendix~\ref{app:cases}.

\begin{figure}[!tb]
\centering
\includegraphics[width=\textwidth]{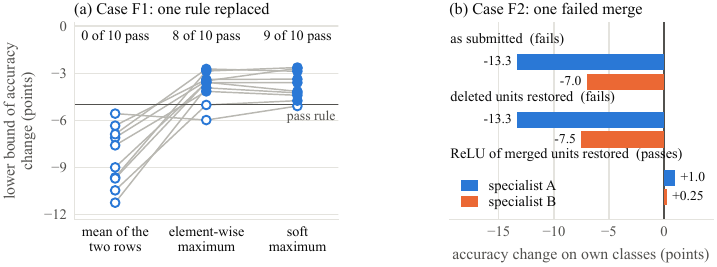}
\caption{Two single-change reruns, holding the networks, hidden data, and grading fixed. (a) Case~F1: the rule that joins two breeds into one class is replaced, on ten instances; each line is one network, a filled point passes, and a network passes if the lower end of the interval of its accuracy change stays above $-5$ points. (b) Case~F2, one failed merge: restoring the ReLU of the units that \astra{} had merged into linear maps removes the loss of accuracy, while restoring the units that it had deleted does not.}
\label{fig:mechanism}
\end{figure}

\textbf{Averaging where the maximum is needed (Case~F1).} Without any photograph, \astra{} must rewrite the final layer of a 20-breed classifier so that it predicts ten new classes, each of which joins two look-alike breeds. In all ten instances of this task, \astra{} submitted, without a test, a layer in which each new class is the average of the two breeds' weight rows or a blend close to it, and none of these layers passed. The average is the wrong rule because a photograph shows only one of the two breeds, and the averaged row gives it a lower score than the row of that breed alone, so another class can win (Result~\ref{r:fail}). Taking the element-wise maximum of the two weight rows instead passes on eight of the ten networks (Figure~\ref{fig:mechanism}a). When we added a single factual hint to the task statement, namely that the features entering the final layer are never negative, \astra{} found this maximum rule by itself and passed eight of ten instances.

\textbf{An assumption that was never checked (Case~F2).} \astra{} must merge two specialist classifiers into one network of the same size. In each specialist, only half of the retrained units work on photographs, and in the harder versions the idle half was trained to respond to artificial images, so that it looks active. In the failed attempts, \astra{} chose which units to keep by their activity on images that it had made itself, on which the idle units look important. It then merged units on the unchecked assumption that their ReLU never cuts anything off, and it tested the merged network on the same artificial images, where the error does not show (Figure~\ref{fig:mechanism}b). As the training of the idle units grew stronger over the four versions, the passes fell from 10 to 7, 3 and 2 of the 10 attempts (9 of 9, 7 of 10, 2 of 9 and 2 of 10 scored instances; test for trend, $p<0.001$ for both counts). Yet two rules of 15 and 30 lines, which use only the weights and keep the units that the retraining changed most, solve 29 of the 30 attempts across the three versions with misleading idle units.

\textbf{Other failures (Cases~F3 to~F8).} The other failures that we analyzed show the same pattern (Appendix~\ref{app:cases}). In the family with the most failures, 31 of the 43 scored instances failed, and in all 43 \astra{} chose the size of a repair by judgment and never ran the repaired network. In a compression task, it checked its compressed encoder only on the 64 examples from which it had built it. In a task requiring \astra{} to name the outputs of a classifier, it matched breeds to outputs by the similarity of their weights alone, so a single swapped pair of look-alike breeds was enough to fail. For several of these families, a short program of 2 to about 40 lines, given in the appendix, solves many of the instances. \astra{}'s difficulty therefore seems to lie less in writing the code than in choosing the right approach and checking the assumptions behind it.

\subsection{Other language models}
\label{sec:models}

\begin{figure}[!tb]
\centering
\includegraphics[width=\textwidth]{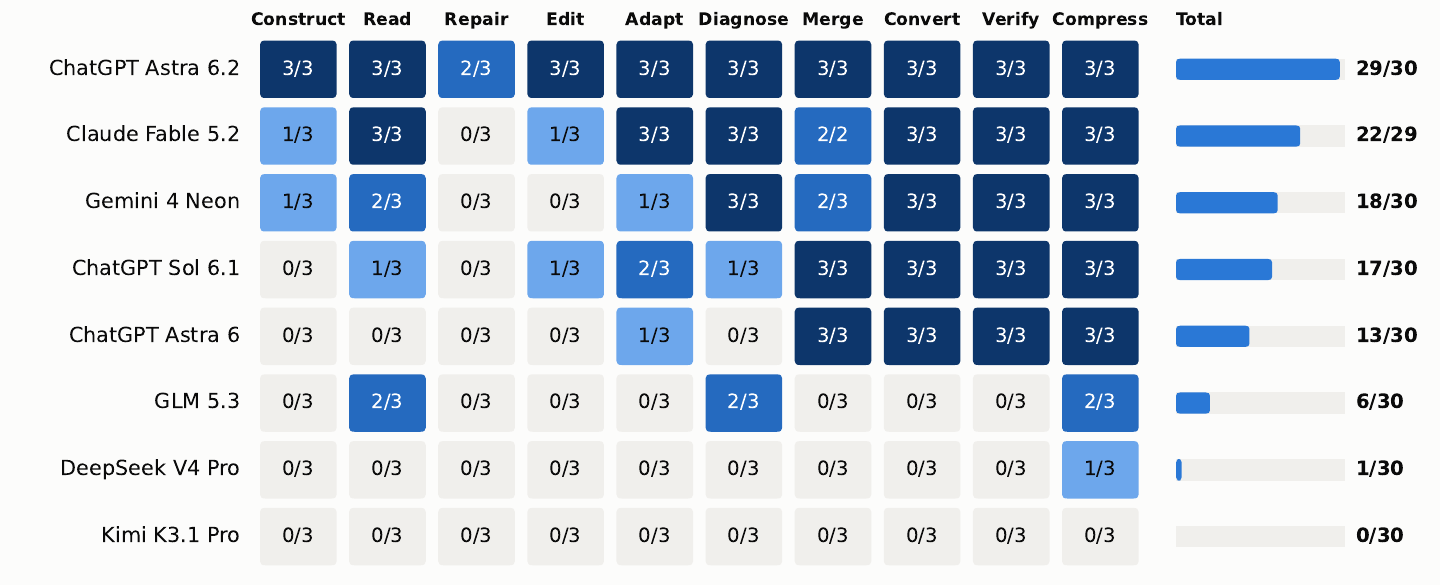}
\caption{Instances passed by eight language models on ten tasks of \web{} (one per operation, three instances each) after rule-compliance auditing, ranked by total passes.}
\label{fig:models}
\end{figure}

To see whether these limitations are specific to \astra{} or shared across models, we evaluated seven further language models on 30 instances across ten tasks (one per operation, taking the three instances where \astra{} performed worst). All models received the same task statements and run limits, and every pass underwent the same rule-compliance audit (counting 17 rule-breaking or tool-use sessions as failures). The most striking result is the sharp jump from \astra{} (43\%) to \astra{} 6.2 (97\%, solving 16 of the 17 instances \astra{} failed), pointing to rapid progress in weight-space programming and confirming that all ten tasks are solvable within the protocol. Claude Fable 5.2 (76\%, or 73\% if one borderline Merge session is counted as a failure), Gemini 4 Neon (60\%), and ChatGPT Sol 6.1 (57\%) followed behind \astra{} 6.2, while all three open-weight models trailed the commercial models at 20\% or below (Figure~\ref{fig:models}).

Across task types, the five commercial models mirror the main benchmark's divide: every commercial model solved all nine constructed-weight instances (whereas the three open-weight models passed only 3 of 27 combined), while unseen photographs separated the commercial models sharply. On those 15 photograph instances, \astra{} passed 1, ChatGPT Sol 6.1 and Gemini 4 Neon passed 4 each, Claude Fable 5.2 passed 8, and \astra{} 6.2 passed 14, with repairing wrongly labeled classifiers remaining the hardest task (solved only by \astra{} 6.2; Section~\ref{sec:limits} notes the sample and setup caveats of this comparison).

\subsection{Exploits of the task files}
\label{sec:exploits}

In early versions of a few tasks, some of \astra{}'s most elaborate programs bypassed the intended problem by exploiting artifacts left behind when the task files were generated. We excluded all instances solved this way (condition~(b) of Definition~2) and rebuilt the affected tasks. In early builds of five families, initial weights were drawn from a Mersenne Twister pseudo-random generator with a 32-bit seed~\citep{mt19937}, and training left some of those weights untouched. \astra{} attempted to recover the seed on 25 instances and succeeded on 15. In one instance, it solved a rank-19{,}968 linear system over bits for the generator's full internal state (Result~\ref{r:seed}), regenerated the initial weights, and read off the hidden answers. After the rebuild, the task of naming shuffled classifier outputs dropped from four of five passes to zero of five. Appendix~\ref{app:leakmath} documents a second exploit, in a verification task whose candidate repairs were saved in higher numerical precision than the hidden network, along with the mathematics of both exploits and related work.

\section{Discussion}
\label{sec:discussion}

\textbf{The successes can be explained.} Modifying an unfamiliar network directly through its weights, without data or retraining, can look mysterious. Yet the successful programs we reviewed reduce to established mathematics: the Chinese remainder theorem, Richardson extrapolation, network symmetries, linear systems for input covariances, and Gaussian models of hidden units. Some of these methods are exact, while others rely on statistical assumptions tested on held-out data. Our contribution is therefore not new mathematics, but the finding that a language model, given only the weights and a task statement, can recognize which methods apply and combine them unprompted. Even where a task statement names a structural cue (such as symmetry in Case~S7, Gaussian inputs in Case~S6, or non-negative weights in Case~S8), \astra{} still had to derive and implement the full procedure.

\textbf{What generalizes, and what does not.} Across \web{}, \astra{} passed every scored instance on constructed weights in all ten operations, including rare activation functions and million-step recurrences. Many of the successful techniques we examined share three traits. First, everything needed to determine the answer lives in the weights or in the specification. Second, the method is exact or carries an error bound over the full input domain, so it does not rely on sampling inputs. Third, assumptions are verified before being applied: \astra{} first proves, for instance, that two counters never coincide or that a unit never clips, and only then builds on that fact. Because an exact rewrite matches the original network on every input, including unseen ones, these methods proved reliable on the networks we tested.

\textbf{Where the failures occur.} Failures cluster where weights were learned from data and grading depends on unseen real data. Even there, \astra{} passed 69\% of the instances, in several families by extracting training-data statistics stored inside the weights (such as normalization variances). The failures we analyzed follow a clear two-part pattern: \astra{} first relied on a flawed assumption or heuristic (such as averaging two classes or assuming units never clip), and then either ran no check or tested only on synthetic inputs or the same few examples used to build the edit. Because short programs of 2 to about 40 lines pass many of the instances across several failing families, \astra{}'s bottleneck appears to lie less in coding than in choosing the right method and verifying its assumptions. These explanations hold for the cases we analyzed in detail and not necessarily for every failure.

\textbf{What to test next.} Our findings point to three concrete questions that can be tested on fresh networks under the same protocol:
\begin{enumerate}
\item Does a solver improve if it must propose and compare multiple candidate methods before committing to one?
\item Can solvers be equipped with self-checks that catch errors missed by synthetic probes or by reusing the design examples?
\item Do these failures disappear when a solver is required to verify each mathematical assumption (such as non-negative features or non-clipping units) before simplifying a network?
\end{enumerate}

\section{Limitations}
\label{sec:limits}

\textbf{Model coverage and the multi-model comparison.} We evaluated \astra{} across the full benchmark and seven additional models on 30 instances across ten tasks (Section~\ref{sec:models}), all under one protocol that forbids fitting and allows at most 50 runs. Two caveats apply to that 30-instance comparison. First, because the instances were chosen where \astra{} performed worst (without rerunning \astra{}, which passed 59\% across all scored instances of those ten tasks), regression to the mean accounts for part of the apparent gain and smaller pairwise gaps are not statistically distinguishable. Second, provider wrappers and reasoning effort differed across models: Gemini 4 Neon ran through Google's Antigravity CLI, Claude Fable 5.2 ran through Claude Code without tools, the open-weight models ran at the maximum reasoning effort available on DeepInfra (which may not match the effort levels of the commercial models), and Kimi K3.1 Pro failed to follow the harness action format, confounding the gap between commercial and open-weight models.

\textbf{Benchmark scope and network scale.} Because we added task families while investigating \astra{}'s failures, pass rates characterize this curated benchmark rather than a representative sample of all weight-engineering problems. Our networks are also small by design so that failures cannot be attributed to scale: learned networks have at most about 400{,}000 parameters and the largest constructed network has about 2.4 million. None of the target networks is itself a large language model, so these results do not establish how weight-space programming scales to large models. In addition, the two failure-associated features of Section~\ref{sec:contrasts} cannot be fully separated, because every unseen-data task that supplies weights uses learned weights.

\textbf{Scoring rules and automated auditing.} A miss is scored only when a permitted solution has been demonstrated for that instance, whereas a compliant pass establishes feasibility directly. If the 149 misses of the retained tasks without such a demonstration were counted as failures, the overall pass rate would fall from 83\% to 76\%. Finally, language models carried out the bulk of the compliance audits and method annotations. In particular, Claude Fable 5.2 helped draft the task statements, generators, and graders and annotated the benchmark records before being evaluated in Section~\ref{sec:models}, and the automated compliance reviewer (Claude Opus 5.5) comes from the same developer.

\clearpage
\normalsize
\appendix
\providecommand{\qed}{\hfill$\square$}
\newcounter{wres}
\renewcommand{\thewres}{\thesection\arabic{wres}}
\newenvironment{wres}[1][]{\refstepcounter{wres}\par\smallskip\noindent\textbf{Result~\thewres\if\relax\detokenize{#1}\relax\else\ (#1)\fi.}\ }{\par}
\counterwithin{table}{section}
\counterwithin{figure}{section}

\section{Case studies}
\label{app:cases}
This appendix gives the full record behind the techniques and the failures of Sections~\ref{sec:success} and~\ref{sec:contrasts}. Each case lists its facts (task, network, what the solver was given, hidden test, pass rule, what the answer depends on) and then summarizes the solver's approach and outcome. Cases~S1 to~S16 are mainly successes (Cases~S12 and~S16 include failures) and Cases~F1 to~F8 are failures. A ``test for trend'' is a Cochran--Armitage test over all instances that were attempted at each level of a ladder; it is descriptive, because the levels were built while we observed the model. Instances are scored by the rule of Section~\ref{sec:def}; comparisons between versions of a family count all instances that were attempted, and each case says how many are scored. A pass rule ``with a 95\% margin'' requires the unfavorable end of a 95\% confidence interval appropriate to the task's metric (such as a two-sided Wilson interval for proportions, or normal, quantile, and ratio intervals for continuous metrics), rather than the raw point estimate, to meet the threshold. ``Own images'' are images that the solver generates itself.

\textbf{Further techniques.} Several other successful techniques can also be explained mathematically, and the cases below describe them. \astra{} reproduced a counting network's outputs with a GRU for up to a million steps, using multiscale error cancellation in three instances, as in Richardson extrapolation~\citep{richardson} (40 of 40 instances; Results~\ref{r:lanes} to~\ref{r:exact}). It repaired overwritten units of a network from the symmetry of its weights, without being told which units were damaged (34 of 34; Result~\ref{r:twin}). It also merged two recurrent networks through a subspace that they share (Result~\ref{r:gru}), proved the largest output of a six-layer network in exact fractions (Result~\ref{r:cert}), and recovered the mean and covariance of a network's inputs from its stored statistics (Result~\ref{r:price}). Appendix~\ref{app:proofs} gives the mathematics behind the main techniques, with worked examples.

\textbf{Annotation.} The records of 2{,}088 instances were annotated at three levels by two models, Claude Fable 5.2 and ChatGPT Sol 6.1 (\texttt{gpt-6.1-sol}), a different model from the evaluated GPT-6-Astra. The first worked as follows. At the first level a reader (Claude Fable 5.2, working through a batch of instances) read every program that \astra{} wrote for an instance in full, worked out from the code what it does, and wrote a record in a fixed format, including a summary, the methods used, how the answer was tested, any technique that a skilled engineer would not expect, and, for a miss, its direct cause and the decision behind it. The reader formed its view before reading the note of the automated reviewer of that record (Claude Opus 5.5, which did not know the result) and recorded where it disagreed, and a program checked the format of every record. The reader saw the result. The second level summarizes the records of each task family and the third those of each operation. The labels are judgments made instance by instance, so counts of labels show the size of a pattern and are not measurements. ChatGPT Sol 6.1 analyzed the records at the same three levels and added derivations, reruns of failed instances and a synthesis; we recomputed its numbers, repeated its reruns and checked its derivations before using them.

\textbf{Checks and labels.} Among the scored instances whose answer depends on unseen data, by the kind of check that the annotation records, 53 of 119 instances passed with no check, 161 of 244 with a check by computation only and 296 of 369 with runs of a network. These are associations across different tasks and not measured effects of running the network. Without a run, \astra{} can still check its answer by computation on the weights. According to the annotation, it made such a check on 83 of the 157 annotated instances among the 163 failures without a run and none on 74; a check of this kind tests an answer against the assumptions behind it and cannot show that the assumptions hold on unseen data. The annotation gave each failed instance one to five labels. Among the 235 annotated failures the most frequent labels were an answer chosen on weak evidence (146), a wrong rule for combining weights (71), a missing idea (62), a self-check that was trusted too much (60), a check on inputs that could not show the error (41) and a design fitted to a few examples (39). The labels overlap and are judgments of the annotating models.

\textbf{The three groups coded differently.} The three groups of Table~\ref{tab:ops} are coded from the task descriptions, and two choices could be made differently. Three families that we count under unseen data, because they are graded on hidden photographs, are solved by rewriting the units, with no use of the photographs (132 instances, all passed; Cases~S8 and~S10). Only one of the three rewrites is exact for every input, the merge proved from the signs of the weights (70 instances); the rewrite with absolute-value units was checked on supplied example inputs (52), and the rewrite with wave units is an approximation by a Fourier series (10). Three Diagnose families that we count under a stated model ask for a fact of the training history on photographs (112 instances, 99 passed). Section~\ref{sec:contrasts} gives the counts with both recoded.

\textbf{Groups of techniques.} The annotation records the methods that the programs written for each instance use (Section~\ref{sec:success}). The annotations were summarized for each task family and then for each operation, which gave 199 named techniques, and these were sorted by hand into the 14 groups of Table~\ref{tab:matrix} in the main text; another sorting would move some cells. A cell is filled when at least one technique of the group was used in the operation. Several techniques can be used on one instance, so the numbers of a row or column do not add up to numbers of instances, and no cell isolates the effect of a technique from the rest of the program. Table~\ref{tab:tech} lists the most striking techniques.

\begin{table}[!ht]
\centering\footnotesize
\caption{The most striking techniques. \emph{Used}: instances on which the programs use the technique, among all instances that were attempted, with those that passed in parentheses; the counts come from the annotation of the programs and, for rows marked $^\ast$, were confirmed by a text search for the formula or function call that the technique needs. \emph{Scored}: passed/scored instances of the families concerned. A ReLU unit \emph{clips} when its input is negative; a unit that never clips acts as a linear map.}
\label{tab:tech}
\setlength{\tabcolsep}{3.5pt}
\begin{tabularx}{\textwidth}{@{}Xlll@{}}
\toprule
Technique (operations) & Used (passed) & Scored & Case\\
\midrule
Covariance of unseen inputs from stored normalization statistics; best linear fill-in, Prop.~1 (Adapt, Read, Compress) & 59 (49), 20 (20), 20 (20) & 49/50, 20/20, 20/20 & S1, S6\\
Hermite series for rectified channels; next layer by least squares, Prop.~2 (Compress) & 42 (42)$^\ast$ & 50/50 & S2\\
Congruences and the Chinese remainder theorem for networks of counters (Construct, Read, Compress, Convert) & 102 (102)$^\ast$ & 95/95 & S3\\
Counter in a GRU with errors canceled across scales (Convert) & 3 (3) & 40/40 & S4\\
One reachable subspace shared by two recurrent networks (Merge) & 70 (70) & 70/70 & S5\\
Sine and snake units made to pass a signal unchanged; absolute-value units that carry linear signals (Convert) & 49 (48); 96 (95) & 40/40; 52/52 & S10\\
Overwritten units rebuilt from the symmetry group of the network (Repair) & 40 (40)$^\ast$ & 34/34 & S7\\
ReLU units proved never to clip, merged into linear maps (Merge) & 90 (90)$^\ast$ & 70/70 & S8\\
Exact proofs of maxima, of the equality or difference of two networks, and of removable units (Read, Verify, Diagnose) & 20 (20), 30 (30), 15 (15) & 20/20, 15/15, 9/9 & S9, S14\\
\bottomrule
\end{tabularx}
\end{table}

\casehead{Case S1 (Adapt). Stored data statistics: color values deleted from every photograph}
\casefacts{Adapt a classifier of 12 cat and dog breeds to photographs from which color values were deleted: only $r$ of the 12 color values in each $2\times2$ block of pixels remain ($r$ from 9 down to 5). The adapted classifier must give the answer that the original gives on the complete photograph.}
{Convolutional classifier of $64\times64$ color photographs: 96 first-layer filters, each reading one $2\times2$ block (12 color values), five more convolutional and two fully connected layers, batch normalization~\citep{bn} after each convolution. 390{,}668 numbers (weights and stored statistics); trained on 1{,}560 photographs, 130 per breed.}
{All weights, including the mean and variance stored by each normalization layer; the list of the values that remain. No photographs. Up to 50 runs on at most 16 own images.}
{600 photographs, 50 per breed, with the same values deleted.}
{Same breed as the original on at least 413 of the 600 (68.8\%): true agreement above 65\% with a 95\% margin.}
{Unseen data: real photographs.}

\emph{What the solver did.} The missing color values can only be estimated from the statistics of real photographs, and no photograph is supplied. \astra{} noticed that the network itself stores such statistics. Stored normalization statistics have been used before to calibrate or distill a network without data~\citep{dfq,deepinv}; here they are solved for the covariance of the input. Each of the 96 filters of the first layer computes a weighted sum $a_k^{\top}x$ of the 12 color values $x$ of a block, and the normalization layer stores the mean and the variance of that sum over the training photographs.

\textbf{Proposition 1 (data statistics from the weights).} Let the input $x\in\R^{d}$ have mean $m$ and covariance $C$ on the training data, and let unit $k$ of the first layer compute $a_k^{\top}x+b_k$ and store its mean $\mu_k$ and its variance $\sigma_k^{2}$. Then $\mu_k=a_k^{\top}m+b_k$ is linear in $m$, and $\sigma_k^{2}=a_k^{\top}Ca_k$ is linear in the $d(d+1)/2$ entries of $C$, so $K$ units determine $C$ whenever the $K\times d(d+1)/2$ coefficient matrix has full column rank. If only $y=Nx$ is observed and $NCN^{\top}$ is invertible, the reconstruction $\hat x=m+L(y-Nm)$ has the smallest mean squared error among affine estimators for $L^{\star}=CN^{\top}(NCN^{\top})^{-1}$.

Here $d=12$, so $C$ has 78 unknown entries, and the 96 stored variances give 96 equations of rank 78. We checked this on all ten networks of the family, which are the same at every level: the coefficient matrix has rank 78 on each, and the ratio of its largest to its smallest singular value lies between 27 and 39, so the solution is well conditioned. \astra{} solved them, computed the fill-in $L^{\star}$ for the values that remain and wrote it into the weights of the first layer, so that the rest of the network receives estimates of all 12 color values. The stored statistics were accumulated during training, so the equations hold to the accuracy of that record.

\emph{Result.} 49 of the 50 instances passed, 19 without a run.

\casehead{Case S2 (Compress). Gaussian models of units: pruning a photo classifier without a single photograph}
\casefacts{Replace the 128 channels of the last convolutional layer of a classifier of 37 cat and dog breeds by 12 to 28 channels, and rewrite the next layer to read the smaller layer. No other layer changes.}
{Convolutional classifier of $64\times64$ color photographs: six convolutional layers, each followed by batch normalization, and two fully connected layers. 37 outputs; 332{,}901 numbers (weights and stored statistics).}
{All weights and the stored mean and variance of every channel. No photographs. Up to 50 runs on at most 16 own images. Training the smaller network to imitate the original is forbidden.}
{1{,}848 photographs of the 37 breeds, not used in training.}
{Accuracy at most 5 percentage points below the original network, with a 95\% margin.}
{Unseen data: real photographs.}

\emph{What the solver did.} To decide which channels can be merged, one must know which of them respond together on real photographs, and the file stores only the mean and the variance of each channel separately. \astra{} modeled the signals before the ReLU as jointly Gaussian, with means and variances taken from the stored statistics. It estimated the correlation between two channels from the inner product of their filters, as if the signals of the layer below were uncorrelated, with one fitted number for the correlation between neighboring positions. It then computed the covariance of the rectified channels in closed form.

\textbf{Proposition 2 (covariance of rectified units; \citealp{wright}).} Let $(U,V)$ be jointly normal with standard margins and correlation $\rho$, and let $\phi$, $\Phi$ and $\He_n$ denote the standard normal density, its distribution function and the probabilists' Hermite polynomials ($\Phi$ is not the grader of Definition~1 here). Then
\begin{equation}
\begin{aligned}
\Cov\big(\relu(U+s),\relu(V+t)\big)&=\sum_{n\ge1}c_n(s)\,c_n(t)\,\rho^{n},\\
c_1(t)=\Phi(t),\qquad c_n(t)&=\phi(t)\,\He_{n-2}(-t)/\sqrt{n!}\quad(n\ge2).
\end{aligned}
\label{eq:hermite}
\end{equation}

The series is a published result, Theorem~1 of \citet{wright}
for ReLU units, and
the corresponding uncentered ReLU moment is the arc-cosine kernel
of \citet{chosaul};
the covariance is obtained by subtracting the product of the means,
and what we report is its use. Each new channel is a mixture of old channels with a new threshold. \astra{} chose the mixtures by a numerical search that minimizes, under the model, the error with which the old features can be rebuilt linearly from the new channels. The weights of the next layer then follow by least squares, $X=K_{xs}K_{ss}^{-1}$, where $K_{ss}$ is the model covariance of the new features and $K_{xs}$ that between old and new features. The search uses the model and never measured behavior, so it is not fitting in the sense of Section~\ref{sec:def}.

\begin{figure}[!ht]
\centering
\includegraphics[width=\textwidth]{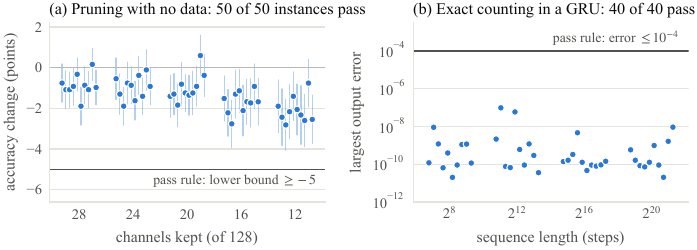}
\caption{Two derived solutions, measured on hidden tests; every point is one instance. (a) Case~S2: change in the accuracy of a photo classifier after its last convolutional layer was pruned without any data, with 95\% intervals; an instance passes if the lower end stays above $-5$ points. (b) Case~S4: largest output error of a GRU that replaces a constructed counting network; an instance passes if the error stays below $10^{-4}$.}
\label{fig:success}
\end{figure}

\emph{Result.} All 50 instances passed (Figure~\ref{fig:success}a); on 42 \astra{} used this method, and on 26 it made no run. The mean change of accuracy was $-0.9$ points with 28 channels and $-2.1$ points with 12. The Gaussian model is an assumption that only the hidden test checks; it worked where the statistics are stored next to the layer that changes, and never where \astra{} invented them or pushed the model through many layers (0 of 39 instances that were attempted). On three instances with 12 channels it used a variant in which the compressed signals are shifted so far that the following ReLU never cuts them off (Result~\ref{r:shift}); all three passed.

\casehead{Case S3 (Read). Modular arithmetic: cycle lengths of a network of counters}
\casefacts{Give the exact cycle length of each of the four outputs of a recurrent network that runs without input. The cycles can be far longer than any run.}
{Constructed recurrent ReLU network with 23 units and four outputs: $h_{t+1}=\relu(Wh_t+b)$, $y_t=Vh_t+c$. About 650 numbers, all integers times powers of two, so that it runs without rounding.}
{All weights and the starting state. Up to 50 runs of 1{,}024 steps each.}
{The true cycle lengths, computed by the builders.}
{All four numbers exactly right.}
{The weights alone.}

\emph{What the solver did.} \astra{} rescaled the units by positive constants, which does not change the function of a ReLU network, until every weight was an integer. Pairs of units then appear as counters that count modulo some number $m$ and emit a carry, and detector units fire when several counters are in given states at the same step. Each firing time is thus a system of congruences $t\equiv a_i\pmod{m_i}$. By the Chinese remainder theorem, two events $t\equiv a\pmod m$ and $t\equiv b\pmod n$ coincide if and only if $\gcd(m,n)$ divides $a-b$, and then exactly once every $\operatorname{lcm}(m,n)$ steps. If it does not divide, the detector never fires, which is a proof and not an observation. A counter that is driven by such an event multiplies its period. In one network two clocks with periods 152 and 428 coincide every $\operatorname{lcm}(152,428)=16{,}264$ steps, this event drives a chain of four counters that divide by 488, 234, 66 and 220, and the last output therefore has the period $16{,}264\times488\times234\times66\times220=26{,}966{,}815{,}349{,}760$ (Result~\ref{r:clock}).

\emph{Result.} All 20 instances passed, and \astra{} ran the network on none of them. The same arithmetic solved three further families, in which such networks are built, shrunk and made to run faster. All 95 scored instances of the four families passed, and the programs of 65 of them call routines for greatest common divisors, least common multiples or the remainder theorem.

\casehead{Case S4 (Convert). Error cancellation across scales: a counting network rewritten as a GRU}
\casefacts{Write a GRU~\citep{gru} that reproduces the four outputs of a given recurrent ReLU network at every step of sequences of up to 1{,}048{,}576 steps. A GRU is a recurrent network whose memory passes through $\tanh$ and whose \emph{gates}, units with values between 0 and 1, decide how much of the old memory is kept.}
{Source: constructed recurrent ReLU network with 40 to 52 units, 3 inputs and 4 outputs, containing a small state machine, two chained counters that count modulo numbers of up to a few hundred, a delay line and memory cells; its units are shuffled and rescaled. Target: a GRU with 41 to 52 units (about 8{,}000 parameters).}
{The weights of the source, a short description of the kinds of parts inside it, and the rule that generates the test sequences. Up to 50 runs on sequences of at most 4{,}096 steps.}
{16 sequences of the stated length.}
{Each output within $10^{-4}$ of the source at every step.}
{The weights alone.}

\emph{What the solver did.} A count that is stored in a GRU passes through $\tanh$ at every step. Because $\tanh$ bends, the stored count drifts, and finite-precision GRUs cannot count without bound~\citep{weiss}, although networks of sigmoidal units with unbounded precision can simulate any machine~\citep{kilian}. Here the count is bounded and the length is fixed, and \astra{} found five designs that all passed. It first recovered the parts of the source from its weights and built the logic from gates driven so hard that, in double-precision arithmetic, they are exactly 0 or 1. For the counters it used digital counters built from such gates (17 instances), a unit that stores the count at a tiny scale, where $\tanh$ is linear to within a proven bound, a device also used in approximation results for $\tanh$ networks~\citep{deryck} (11), a unit that resets itself in every cycle (8), a geometric code (1), and a count stored at several scales at once (3). In this last design the count $u$ is stored as $\tanh(s_iu)$ at $k$ small scales $s_i$ and read back as $F(u)=\sum_i\alpha_i\tanh(s_iu)/s_i$ with $\alpha_i=\prod_{j\ne i}s_j^2/(s_j^2-s_i^2)$. These weights cancel every error term up to order $2k-1$, as in Richardson extrapolation~\citep{richardson} (Results~\ref{r:lanes} to~\ref{r:exact}). With six scales from $2^{-8}$ to $2^{-13}$ and counts up to 139, the error per step is below $2.7\times10^{-13}$ and the accumulated error over $2^{20}$ steps is below $3\times10^{-7}$, before rounding.

\emph{Result.} All 40 instances passed, with a largest output error of at most $9.6\times10^{-8}$ across all lengths (Figure~\ref{fig:success}b). Because a permitted run covers only 4{,}096 of the million steps, \astra{} relied on a leading-term error estimate rather than a full simulation, and Result~\ref{r:exact} supplies the complete remainder and non-expansion bound.

\casehead{Case S5 (Merge). Two recurrent networks merged through the memory that they share}
\casefacts{Build one GRU with 44 to 80 units that gives, at every step, the four outputs of each of two GRUs that read the same input sequence.}
{Two GRUs with 64 units, 8 input channels and 4 outputs each (14{,}468 parameters each), which predict quantities of one simulated dynamical system. In each, about 16 units were trained, and about 48 form a block with nearly constant gates and small signals, built so that the two blocks carry the same underlying signals. The task statement does not disclose this. A side-by-side copy would need 128 units.}
{All weights of both networks and the rule that generates the inputs. Up to 50 runs, which also show the memory and the gate values.}
{256 input sequences of 256 steps.}
{For each source, root-mean-square error at most 5\% of the standard deviation of that source's outputs, over the whole sequence and over its last quarter, with a 95\% margin.}
{A statistic of the weights: the test sequences follow a stated rule.}

\emph{What the solver did.} Reading the weights, \astra{} saw that about 48 of the 64 units of each network have nearly constant gates and tiny signals, so that they update almost linearly. Two networks that read the same inputs form one stacked linear system $x_{t+1}=Fx_t+Gu_t+d$ with $x_0=0$, and the state of such a system can only reach the span of the columns of $L$, $FL$, $F^2L,\dots$ with $L=[G\ d]$. For the nearly linear units of both networks together, 96 in the instance that we examined, this reachable subspace has only 15 or 16 dimensions. \astra{} built this small system a single time inside the merged GRU, in $\tanh$ units at a signal scale so small that $\tanh$ is linear, read the nearly linear units of both networks from it, and copied the remaining units of each network (Result~\ref{r:gru}). This needs about $16+16+16=48$ units; for the limit of 44 it cut the shared system to its 12 most important directions.

\emph{Result.} All 70 instances with 44 to 80 units passed. The reduction is the reachable-subspace argument of linear systems theory~\citep{kalman}; a related reduction of recurrent networks is canonicalization~\citep{grigoryeva}. It is exact for the linear model of these units, which are linear only approximately, and the hidden test validates the whole network.

\casehead{Case S6 (Read). The input distribution behind stored statistics}
\casefacts{Recover the mean of each of 12 inputs and the covariance of every pair of inputs, from a network that stores statistics of its internal signals.}
{A network with 12 inputs, 48 to 312 hidden units and 96 outputs, with a normalization step after each of its two layers (about 5{,}900 numbers in the smallest version). The task says that the inputs are Gaussian and that the stored means and variances are exact for that distribution; it gives neither the mean nor the covariance.}
{All weights and the stored statistics. Up to 50 runs, which reveal nothing that the stored numbers do not contain.}
{The mean vector and the covariance matrix from which the statistics were computed.}
{Every submitted number within 0.001 of the true value.}
{A statistic of the weights, under the stated Gaussian model.}
All 20 instances passed without a run. In three versions the first layer alone determines the covariance (Proposition~1). In the hardest version each first-layer unit reads only a few inputs, which leaves at least 56 of the 78 covariance entries open. \astra{} then used the 96 variances stored in the second layer. Each is a known function of the correlations between rectified Gaussian signals (Result~\ref{r:price}), and solving these equations fixed the open entries. The largest error over the 20 instances is $2.2\times10^{-12}$.

\casehead{Case S7 (Repair). Symmetry: overwritten units rebuilt}
\casefacts{Restore the function of a network in which 4 to 9 hidden units were overwritten with wrong values. Only the damaged units may be changed, and the task does not say which they are.}
{Constructed ReLU network with 12 to 18 inputs, a few dozen hidden units in one layer and one output. Its output does not change when the inputs are reordered in certain stated ways.}
{The damaged weights, the reorderings that leave the original output unchanged, and the number of damaged units. Up to 50 runs.}
{2{,}048 inputs and the original network.}
{Every changed unit is truly damaged, and the output matches the original to within $10^{-6}$ on all hidden inputs.}
{The weights alone.}
All 34 scored instances passed without a run. Because of the symmetry, the healthy units come in groups whose members mirror each other. \astra{} listed the reorderings, sorted the units into such groups, found the units that fit no group and rewrote them as the missing members (Result~\ref{r:twin}). In one instance 8 of 36 units were damaged, the symmetry group had 24 elements and the error was zero on all 2{,}048 inputs.

\casehead{Case S8 (Merge). Signs of the weights: two classifier heads merged}
\casefacts{Write one classifier head with far fewer hidden units that gives the scores of two heads which share one image encoder.}
{A shared encoder turns a photograph into 128 features that are never negative. Head A and head B each map them through 128 hidden ReLU units to the scores of six to nine breeds. The merged head may have 52 to 128 hidden units.}
{All weights. In most versions a sentence says that some hidden units were trained with non-negative weights; two versions give a vaguer hint or none. No photographs. Up to 50 runs.}
{300 to 450 photographs for each head.}
{Agreement with each original head on at least 90\% of that head's photographs, with a 95\% margin.}
{Unseen data: real photographs. An exact rewrite of the heads solves it for every photograph.}
All 70 scored instances passed. \astra{} noticed that the features are never negative and that about 110 of the 128 hidden units of each head have only non-negative weights and offsets. The input of such a unit is never negative, so its ReLU does nothing, and all such units of a head together are one linear map. A linear map to $K$ scores can be carried by $K+1$ ReLU units whose inputs are again never negative (Result~\ref{r:on}). The remaining units were copied. This is the merging step of Case~F2, here preceded by a proof.

\casehead{Case S9 (Read). The largest output of a deep network, with a proof}
\casefacts{Give the largest output that a network can produce when each input lies between $-1$ and $1$, and an input that produces it.}
{Constructed ReLU network with 12 inputs, six or more hidden layers of 64 units and one output. The largest value sits on a narrow spike that random inputs do not find, and the task does not disclose that the spike exists.}
{All weights and the allowed range. Up to 50 runs.}
{The true largest output, which the builders proved for each network.}
{The output at the submitted input and the claimed value are both within $10^{-6}$ of the true maximum.}
{The weights alone.}
All 20 instances passed. \astra{} traced each unit back to the inputs in exact arithmetic, merged units that are copies or mirror images of each other and removed units that can never switch on. The remaining problem was a small linear program, and its solution was confirmed by a dual certificate in exact fractions (Result~\ref{r:cert}). Typically \astra{} used one run, to confirm the result.

\casehead{Case S10 (Convert). Identities between unusual units: a ReLU network rebuilt without ReLU}
\casefacts{Fill in the weights of a new network whose layers are fixed in advance and contain no ReLU, so that it gives the outputs of a given ReLU network everywhere in a box of inputs. The layers use the absolute value, a triangle wave, the sine and similar smooth curves, or multiply pairs of units.}
{Constructed ReLU network with 3 to 5 inputs, two or three hidden layers and 2 to 4 outputs, in which every weight is a simple binary fraction. The new network has 3, 5, 8 or 12 layers of stated types and sizes.}
{All weights, the box of allowed inputs and the layout of the new network. Up to 50 runs.}
{About 3{,}000 inputs in the box: random points, its corners and faces, and points at which a unit of the original switches on or off.}
{Every output within one part in 100{,}000 of the original at every one of these inputs.}
{The weights alone.}
All 20 scored instances passed. \astra{} determined which units of the original network are always on or always off inside the box and rewrote the remaining ReLU units using absolute values, since $\relu(z)=\tfrac12(z+|z|)$ (Result~\ref{r:abs}). It routed signals through the sine and other smooth layers at a tiny scale where they behave linearly, then scaled them back; at scale $2^{-44}$, the error of a chain of three such layers is below $5.7\times10^{-10}$ per carried signal (and at most $2.3\times10^{-9}$ at the output), well within the tolerance of $10^{-5}$ (Result~\ref{r:wire}). Layers that multiply pairs of units acted as pass-throughs via multiplication by 1, and where the final layer computes the maximum of three values minus their mean, \astra{} used that step as the last ReLU (Result~\ref{r:mmm}). Two related families apply the same identities to learned weights. In one, the final part of a photo classifier (128 ReLU units and 12 outputs) must be rebuilt with 84 to 96 absolute-value units; \astra{} identified always-on/always-off units and near-copies, mirror images, or canceling pairs (measured on supplied examples rather than proved; Result~\ref{r:abs}), and all 52 scored instances passed. In the other, each ReLU of a small image classifier must be replaced by 16 or 8 Snake units~\citep{snake}; \astra{} built the replacement from the Fourier series of $|z|$ (Result~\ref{r:cos}), and all 10 instances passed. The written pass rule of this last family could not be met by any answer on its 128 test images; the grader required agreement with the original on at least 126 of them, the ten passes were kept, and the task statement was corrected after the evaluation.

\casehead{Case S11 (Adapt). Hidden integer rows: decimal weights replaced by small whole numbers}
\casefacts{Find a network with only small whole-number weights that computes the same function as a given ReLU network, up to a stated scaling of each output.}
{Constructed ReLU network with a small number of inputs, two hidden layers of at most ten units each and 3 outputs. A whole-number form exists but is hidden: every unit was rescaled by an arbitrary positive factor, and in the harder versions some units were split into several copies.}
{The weights, the scaling factor of each output, the largest whole number allowed and the layer sizes of the answer. No data. Up to 50 runs.}
{4{,}096 inputs, on which the submitted network is run in exact whole-number arithmetic.}
{Every output within $10^{-6}$ of the original after the stated scaling, and every weight a whole number within the allowed range.}
{The weights alone.}
All 20 instances passed. A positive factor can be moved from a ReLU unit into the next layer without changing the function, so each unit has a whole-number pattern that shows in the ratios of its weights, and a ratio that lies close enough to a fraction with a small denominator determines that fraction (Result~\ref{r:rat}). \astra{} read these patterns, merged the split units and moved the remaining factors into the next layer; in the tightest version, with whole numbers up to 7, it searched the few ways of sharing the factors between layers. It then bounded the difference between its network and the original over the whole range of inputs in exact fractions, $4.3\times10^{-10}$ against the limit of $10^{-6}$ in the instance that we examined.

\casehead{Case S12 (Adapt). A correction that is best on average: a sensor changed by unknown amounts}
\casefacts{Adapt a classifier of sensor readings to a new sensor that reports the same cases differently: pairs of the 12 numbers of a reading are rotated and scaled, and all numbers are shifted, by unknown amounts.}
{Classifier of readings of 12 numbers into 8 classes, with learned weights; its first layer stores the mean and the spread of the readings on which it was trained.}
{All weights; the form of the change, that is, which six pairs are rotated and scaled together and the allowed ranges of angle, scale and shift; and 64, 16, 8 or 4 unlabeled readings from the new sensor or, in two versions, a second network that was trained on the new sensor for another task. Up to 50 runs.}
{10{,}000 cases, each with its class, its old reading and its new reading.}
{Accuracy on the new readings at most 5 points below that of the original on the old readings, with a 95\% margin.}
{A statistic of the weights, under the stated model of the change.}
Of the 49 scored instances 47 passed, among them every instance with 64 or 16 readings and every instance in which the second network was supplied. \astra{} compared the spread of the new readings, or the statistics stored in the second network, with the statistics stored in the original classifier, worked out the rotation and the scaling of each pair, and wrote the correction into the first layer without a run. Where the readings leave the change partly open, it used the correction that is best on average over what remains possible (Result~\ref{r:post}). With 8 or 4 readings and no second network, 5 of 20 instances passed, and the 15 misses are not scored, because no solution by permitted means is demonstrated for these instances. On four further instances with 4 readings (the 4 of 800 fresh draws on which a fixed baseline estimator passes), \astra{} passed two instances and failed two, with accuracy losses of 5.9 and 5.7 points at the unfavorable end against the limit of 5. This result shows the limit of the formula. When the readings leave the sign of a rotation open, the average correction no longer undoes the rotation.

\casehead{Case S13 (Construct). A bound on rounding in place of a test: a state machine as a recurrent network}
\casefacts{Write the weights of a recurrent ReLU network that reads a word of bits and outputs 1 if a given state machine ends in an accepting state, and 0 otherwise.}
{No weights are supplied. The answer is a recurrent network of a stated size, from 31 units for a machine with 5 states and words of 8 bits up to a machine with 14 states and words of 16 bits. It sees all bits at once, mixed by a known scrambling.}
{The state machine as tables, the scrambling of the input bits and the exact shape of the network. Up to 50 runs of a draft.}
{Every possible input word, 256 in the smallest version and 65{,}536 in the largest.}
{Output equal to the answer of the machine to within $10^{-7}$ on every word.}
{The weights alone.}
All 25 scored instances passed, as did all 15 of a second family, in which the network may take only one step for each input bit. \astra{} undid the scrambling exactly, stored the bits in a chain of units that shifts by one place per step, and combined the current state with the next bit in pairs of units. In most instances it submitted its network without a run and bounded the rounding error for every input at once (Result~\ref{r:round}). In one instance the bound is $3.55\times10^{-8}$ against the limit of $10^{-7}$, and the error measured over all 65{,}536 words was $1.8\times10^{-11}$.

\casehead{Case S14 (Diagnose, Verify). Exact proofs: units that a network can lose, and networks that agree}
\casefacts{List every hidden unit of a network that can be switched off, alone or together with one other unit, without ever changing the output anywhere in the allowed range of the inputs.}
{Constructed ReLU network with 12 inputs, two hidden layers with 384, 1{,}536 or 3{,}072 units in all, and one output. It contains traps: units that switch on only in about a billionth of the range, and pairs of units that almost cancel each other.}
{All weights, the allowed range of each input and the exact definition of a removable unit. Up to 50 runs.}
{The true set of removable units, which the builders proved for each network.}
{Every entry of the list right.}
{The weights alone.}
All 9 scored instances passed. \astra{} first grouped the units that respond to the same combination of inputs, which made the network small enough for an exact analysis. It then proved with small optimization problems which units can never switch on, found the canceling pairs from matching weights, and kept every unit that puts a kink into the output, because such a unit cannot be removed (Result~\ref{r:kink}). It almost never ran the network. In a related family \astra{} must decide whether two disguised ReLU networks, each with three hidden layers of 256 units, agree on every allowed input, and must supply either a proof that a checker can verify or an input on which they differ. All 15 scored instances passed. \astra{} reduced every unit to an exact expression and merged equal expressions until the two outputs came out identical, as in Result~\ref{r:cert}, or it found an input on which the networks differ.

\casehead{Case S15 (Edit, Repair). Small exact circuits: rules that override a network}
\casefacts{Build a network that follows a list of rules wherever one of them matches and reproduces the original output exactly everywhere else. A rule says that the output must be a given number when certain input bits have certain values, and where several rules match, the first in the list wins.}
{Constructed ReLU network with one hidden layer and one output, whose inputs are strings of 6, 8 or 10 bits. The list has 3, 5 or 8 rules, and the answer has two hidden layers.}
{The weights, the list of rules and the size of the new network. Up to 50 runs.}
{Every possible input string.}
{Output equal to the required value to within $10^{-7}$ on every string.}
{The weights alone.}
All 15 scored instances passed. \astra{} kept the original units and added one detector for each rule, a unit that is 1 exactly on the pattern of the rule and 0 elsewhere; a second layer lets a rule act only when no earlier rule matches and switches the original output off whenever a rule fires (Result~\ref{r:gad}, parts~(i), (ii) and~(v)). The same circuits solved a Repair family, in which one internal feature must be silenced exactly on the inputs for which an odd number of stated tests hold. \astra{} turned each test into a clean 0 or 1, computed whether the number of true tests is odd (part~(iv)) and switched the feature off, and all 25 scored instances passed, with zero error on all 65{,}536 inputs. A circuit of the same kind was also used on photographs, where a unit is repurposed by a rank-one update (part~(vi)) in order to move the prediction for one breed. In that family 19 of 25 scored instances passed, and 14 of the 21 instances on which this circuit was used. The mathematics says exactly what changes, whereas the photographs on which the new unit fires can only be measured on the few examples supplied.

\casehead{Case S16 (Adapt). The mean color of unseen photographs, read from the biases}
\casefacts{Rewrite a classifier of 12 cat and dog breeds so that it still gives its answers when every photograph is changed in a known way: its colors are mixed and its contrast is lowered, and in the harder versions its brightness is bent along a curve, it is blurred, and noise is added.}
{Convolutional classifier of 12 breeds with about 330{,}000 parameters.}
{All weights, a description of the training, and the exact formula and settings of the change. No photographs. Up to 50 runs on at most 16 own images.}
{600 photographs, 50 per breed, changed in the stated way.}
{Same breed as the original gives on the unchanged photograph on at least 471 of the 600 (78.5\%): true agreement above 75\% with a 95\% margin.}
{Unseen data: real photographs.}
Of the 39 scored instances 35 passed; of the ten instances attempted at each of the four levels of change, 10, 10, 8 and 7 passed (test for trend, $p=0.02$). A change of colors can be undone exactly inside the first layer. Blur and noise destroy detail, and the best correction then depends on the statistics of real photographs, which \astra{} read from the weights. The mean color of the training photographs follows from the biases once the normalization is folded into the first layer (Result~\ref{r:bias}). In 13 instances \astra{} also fitted a model of the smoothness of natural images to the condition that all first-layer filters respond about equally strongly, a step for which there is no proof. Among all 40 instances, \astra{} read the statistics of photographs from the weights in either way on 20, and 16 of these passed.

\casehead{Case F1 (Construct). Averaging where the maximum is needed: new class labels}
\casefacts{Rewrite only the final layer of a classifier of 20 cat and dog breeds so that it predicts new classes: ``cat'' and ``dog'' in the easiest version, ten classes that each join two look-alike breeds in the harder one.}
{Convolutional classifier of $64\times64$ color photographs with 328{,}660 parameters. Its final layer maps 128 features, never negative because they follow a ReLU, to one score per class. The answer is a new final layer: 128 weights and one offset per new class.}
{All weights, the breed of each of the 20 outputs and the list of new classes. No photographs. Up to 50 runs of the classifier, or of a draft of the new layer, on at most 16 own images.}
{1{,}000 photographs of the 20 breeds, and a reference layer of the same shape trained on labeled photographs.}
{Accuracy at most 5 percentage points below that of the reference layer, with a 95\% margin.}
{Unseen data: real photographs.}

For ``cat'' against ``dog'', 7 of 10 instances passed. For the ten look-alike pairs, none of 10 passed; nine are scored failures, and for the tenth network no solution by permitted means is demonstrated. Two further scored versions ask for long-haired against short-haired breeds (5 of 10 passed; seven are scored, five passes and two failures) and for species together with coat length (2 of 10 passed; eight are scored, two passes and six failures). A version that adds a new breed to the 20 was passed in none of 10 instances, and none is scored, because no solution is demonstrated. The family thus has 34 scored instances, 14 passes and 20 failures. In six of the ten instances with look-alike pairs \astra{} gave each new class the average of the rows of its two breeds, and in the other four a blend dominated by the average. It ran its layer on no image on any of the 80 instances that were attempted in this family.

\emph{Why it failed.} A photograph of a pair shows one of the two breeds, so the new class should score as high as the stronger of the two old classes. For non-negative features $h$ and weight rows $v_a,v_b$, the averaged row falls short of the stronger class by exactly $\tfrac12|v_a^{\top}h-v_b^{\top}h|$, whereas the element-wise maximum of the two rows never falls short (Result~\ref{r:fail}). The average is therefore weakest on the photographs on which one breed of the pair is recognized most clearly. Between cats and dogs the scores are far apart, and the same loss rarely changes the decision. The statement holds for networks like these ten; writing the row of a new class directly, without training, is otherwise known from weight imprinting~\citep{qi}, which uses examples of the class.

\emph{One-change rerun and hints.} With networks and tests kept and only the rule replaced (Figure~\ref{fig:mechanism}a), the average passes 0 of 10 instances, the element-wise maximum of weights and offsets 8, and a soft maximum, $\tfrac1\tau\log(e^{\tau a}+e^{\tau b})$ for each pair of weights $a,b$, 9; its constant, $1/\tau$ equal to 0.3 times the standard deviation of the weights of the layer (and, for the offsets, of the offsets), was set on 12 other networks, of which the rule passes 11. So does a procedure of two lines, using the element-wise maximum of the weights with the offset $\log(e^{b_1}+e^{b_2})$. When one true sentence was added to the task statement, that the features entering this layer are never negative, \astra{} wrote the element-wise maximum of the two rows itself in all ten instances, in eight of them with the larger of the two offsets, and passed 8 of 10; with a second sentence and with the full rule it passed 8 and 9. These three versions with hints are controls and are not scored. A related family, in which the final layer must send the photographs of five breeds to their look-alike partners, has the same remedy. A soft maximum of the two rows, in at most ten lines, passes all ten instances, where \astra{} passed 3 of 10.

\casehead{Case F2 (Merge). An assumption that was never checked: two specialists merged}
\casefacts{Merge two specialist classifiers into one network of the same size, with 16 outputs, that keeps the accuracy of each specialist on its own breeds.}
{A base classifier trained on 21 breeds. Specialists A and B were made from it by further training of its last three layers, each on eight other breeds, so that only half of the retrained channels and units do the work on photographs; the two working halves fit side by side in one network. In the harder versions the idle halves were also trained to respond to artificial images.}
{All weights of the base network and both specialists, and a description of the training, which says that the training objective had further terms and does not disclose them. No photographs. Up to 50 runs on at most 16 own images; a run also shows internal signals.}
{About 400 photographs for each group of eight breeds.}
{On each group, at most 4 percentage points of accuracy lost against that group's specialist, with a 95\% margin.}
{Unseen data: real photographs.}

Without the misleading training, 10 of 10 instances passed; as it grew stronger, the passes fell to 7, 3 and 2 of 10 (test for trend, $p<0.001$). Of these 40 instances, 38 are scored, 20 passes and 18 failures; the other two attempts passed but broke a rule, by evaluating network layers on synthetic inputs directly in their own code, and are not scored. Where it passed, \astra{} had kept, for each specialist, the channels and units that its training had changed most, unchanged, and placed the two halves side by side. Where it failed, it had judged the role of each unit from its activity on images that it made itself, and in the harder versions these images activate the idle half; that artificial images can mislead about what a unit does on natural images is known from feature visualization~\citep{geirhos}. The same images then served as the final check. On one instance it recorded 36 agreements in 36 test images for a merged network that had lost 13 points of accuracy on photographs.

\emph{Why it failed.} Unable to tell the halves apart, \astra{} compressed both networks on the instances that it failed. It treated some ReLU units as always active and merged them into linear maps, and deleted units that looked always inactive. Replacing a unit $o_j\,\relu(z_j)$ by $o_jz_j$ changes what the unit passes to the next layer by exactly $-o_j\,\relu(-z_j)$, so it is exact only if the input $z_j$ of the unit is never negative. The merged units lie in the last convolutional layer and in the last hidden layer. For the latter the change is that of the scores themselves, because these units feed the output layer directly; for the former, pooling and the last hidden layer act on it, and the one-change rerun below measures its full effect. According to our annotations, 16 of the 18 failures merged or deleted units on such a guess, and 12 chose units from misleading test images. In one failed instance all 83 merged units received negative inputs on photographs; in a one-change rerun, restoring their ReLU turned losses of 13.3 and 7.0 points on the two groups into gains of 1.0 and 0.25 points, whereas restoring the deleted units did not help (Figure~\ref{fig:mechanism}b). The rerun is a diagnosis of the cause and is not offered as a solution of the task. In another family \astra{} proved from the signs of the weights that the units never clip before it merged them, and all 70 scored instances passed (Case~S8).

\emph{Known solution.} Two rules of 15 and 30 lines that read only the weights, and keep for each specialist the units that its training changed most, pass 29 of the 30 instances with misleading training.

\casehead{Case F3 (Compress). A check on the examples used for the design: a digit encoder with a tighter bottleneck}
\casefacts{Replace the first layer of the encoder of a variational autoencoder, a $128\times64$ matrix, by the product of a $128\times r$ and an $r\times64$ matrix with $r=32$ or $r=24$, so that the encoder's outputs and the images rebuilt by the decoder stay close to the original on new images.}
{Variational autoencoder for $8\times8$ images of handwritten digits. The encoder maps the 64 pixels through 128 ReLU units to 8 means and 8 variances of a Gaussian distribution over 8 internal coordinates; the decoder maps 8 numbers through 128 units back to 64 pixels. 19{,}792 parameters; trained on 1{,}133 images of a collection of 1{,}797 digits~\citep{optdigits}, of which 64 others serve as examples and 600 as the hidden test.}
{All weights of the encoder and the decoder, and 64 example images without labels (16 or 4 in two further versions with $r=32$). Up to 50 runs on at most 16 images each.}
{600 images of the same collection, used neither in training nor as examples.}
{Three checks, each with a 95\% margin: the Kullback--Leibler divergence between the old and the new output distribution of the encoder, per internal coordinate, is at most 0.05 on average and at most 0.15 for the worst 5\% of images, and the images rebuilt by the decoder are at most 2\% worse.}
{Unseen data: real images.}

With $r=32$, 9 of 10 instances passed; with $r=24$, on the same networks and the same 64 examples, 1 of 10. Seven of the nine misses are scored failures; for two networks no solution by permitted means is demonstrated. In all nine misses, image reconstruction stayed within tolerance; the encoder's means had shifted, and this shift accounts for 98.7 to 99.6\% of the divergence.

\emph{Why it failed.} \astra{} chose the directions to keep from the 64 examples and checked its answer on the same 64 examples, where the error looked small; on the hidden digits it was 3 to 20 times larger. With $r=32$ there is room for this error, and with $r=24$ there is none. A better source of typical images is the model's own decoder. Decoding random internal coordinates yields the statistics of typical digits. With only 4 examples \astra{} used the decoder in this way and passed 9 of 10; with 16 examples it mostly did not and passed 6 of 10 (on ten networks each, not a statistically significant difference, $p=0.30$). A procedure of 23 lines that takes the image statistics from the decoder passes 8 of the 10 instances with $r=24$.

\casehead{Case F4 (Repair). Judgment in place of measurement: undoing training on wrong labels}
\casefacts{Repair a classifier of 12 breeds whose last two layers were trained further on wrongly named photographs: in two groups of four look-alike breeds, three breeds carried the name of the fourth. The answer is a correction of rank at most 12 to each of the two layers.}
{Convolutional classifier with 327{,}628 parameters. The same ten networks were saved after 0.5, 1, 2, 4 and 8 rounds of the faulty training (one round uses every wrongly named photograph once), so that later versions carry deeper damage.}
{The damaged weights, the table of wrong names and a full description of the faulty training. No photographs and no copy of the healthy layers. Up to 50 runs on at most 16 own images.}
{About 840 photographs, 70 per breed, and the healthy network.}
{At most 8 percentage points less accurate than the healthy network, over all 12 breeds and over the four breeds whose names were not touched, with a 95\% margin.}
{Unseen data: real photographs.}

After half a round of faulty training, 5 of 10 instances passed; after 1, 2, 4 and 8 rounds, 4, 2, 1 and 1 (test for trend, $p=0.01$). Of the 50 instances, 36 are scored, 13 passes and 23 failures; for 14 networks no solution by permitted means is demonstrated. In none of the 50 instances did \astra{} run the repaired network before submitting it. All 37 misses fell short over all breeds, and only 2 on the untouched breeds. After eight rounds the six breeds that had lost their names were, after the repair, on average 29.9 points below the healthy network, while the two breeds that had received their photographs were 8.0 points above.

\emph{Why it failed.} \astra{} applied a single analytic correction directly from the weights. It mainly removes a uniform shift in each row of weights, which is most of the damage early and a shrinking part of it later, when the damage has spread into the hidden units. Without photographs the size of the correction cannot be measured; \astra{} set it by judgment from weight statistics, and it was usually too small. Programs that it wrote for instances that it passed do more. They also raise the output rows of the breeds that had lost their names. Three such programs, run unchanged on all networks, together repair 9, 10, 5, 8 and 3 of the ten networks at the five levels of damage.

\emph{A related family, with the most failures.} A second family poses the same kind of repair after training on wrong names. In each of three groups of four look-alike breeds, the photographs of three breeds carried the name of the fourth, and the last two layers were trained on them for 2, 4, 8 or 12 rounds; here the networks of the four versions are separate networks and not one network at four depths. \astra{} passed 4, 3, 4 and 0 of 10 instances, and 1 of 7 on an earlier build with 12 rounds, of which three are scored. Of the 43 scored instances, 12 passed and 31 failed. \astra{} ran the repaired network on none of the 43, and the annotation marks 22 of the 31 failures as a repair that removes only part of the damage.

\casehead{Case F5 (Verify). The direction of an effect against its size}
\casefacts{Two verification families on photo classifiers. (a) For each of ten proposed edits to the final layer, say whether the edit raises or lowers the recognition of one named breed. (b) For each of ten proposed repairs of a damaged classifier, say whether the repair brings accuracy to within 8 points of the healthy network.}
{(a) A classifier of 20 breeds; each edit is a table of changes to the final layer, in which a part that matters is mixed with larger changes that hardly matter. (b) A damaged classifier as in Case~F4; some candidate repairs are sound, some too weak or too strong, and some carry extra damage.}
{The weights and the ten candidates. No photographs. The solver may run the original network on its own images, but not the edited or repaired networks.}
{(a) 80 to 90 photographs of each breed. (b) About 840 photographs and the healthy network. Every candidate is at least a stated distance (6, 5, 4 or 3 points) from the threshold of 8 points.}
{All ten verdicts right.}
{Unseen data: real photographs.}

In family (a), all 30 scored instances passed. \astra{} read each edit on the few weights that the breed relies on most, which separates the part that matters from the distractions (Result~\ref{r:sign}). A caution applies to this success. In a fourth version of family (a), which is not scored, a simple read-out of two weight statistics, with numbers set on two other networks, gives every verdict on all 12 of its networks, and the automated review marked 12 of the 30 scored passes as possibly resting on regularities in the way the edits were generated. In family (b), 2 of the 40 instances passed. In those 40 instances, 269 of 400 single verdicts were right, and they fell from 85 to 68, 62 and 54 per 100 as the candidates came closer to the threshold. At the closest level, rejecting every repair would have scored 58.

\emph{Why it failed.} The direction of an effect can be read from the sign of a few weights. Whether accuracy stays within a threshold depends on the size of the effect on photographs, and \astra{} judged repairs by surface weight statistics, rejecting sound repairs with visible auxiliary changes and accepting under-scaled ones with smaller weight norms. Two cautions apply. The two families use different networks and edits, so this is not a controlled comparison. Moreover, family (b) is intrinsically difficult: the best known decision rule by permitted means passes only 5, 3, 3 and 1 of the ten instances at the four distances. Of the 40 instances of the corrected versions, 27 are scored, 2 passes and 25 failures; for the other 13 instances no solution by permitted means is known. Feasibility is demonstrated for 11 of the 25 failures by a rule whose thresholds were set on other networks of the family, and for the other 14 only by a classifier of 43 weight statistics that was trained on other networks of the family with their true verdicts, which a solver does not have.

\casehead{Case F6 (Read). Similar weights, different breeds: naming the outputs of a classifier}
\casefacts{Say which of 12 breeds each output of a classifier stands for, after the outputs were put in a random order.}
{Convolutional classifier of 12 cat and dog breeds with 327{,}628 parameters. The breeds form four groups of three look-alike breeds.}
{All weights and the list of breeds. The names of 7, 5, 3 or none of the outputs, depending on the version. No photographs. Up to 50 runs.}
{The breed on which each output was trained.}
{All 12 outputs paired with the right breed.}
{Unseen data: which breed an output responds to on real photographs.}
With seven names given, 7 of 10 instances passed; with five, three and no names, 6, 1 and 1 of 10 (test for trend, $p<0.001$). Three further instances with no names, on networks for which a fixed procedure names all 12 outputs, were all missed. Of these 43 instances, 29 are scored, 15 passes and 14 failures; for the other networks no solution by permitted means is demonstrated. \astra{} never ran the classifier. It grouped outputs by the similarity of their weights and ordered look-alike breeds by faint color cues traced back through the weights. It named 44 of 50, 56 of 70, 60 of 90 and 49 of 120 unnamed outputs correctly, but one swapped pair of look-alike breeds is enough to fail. A procedure of about 40 lines with a table of breed traits passes 9, 7 and 5 of 10 instances with seven, five and three names.

\casehead{Case F7 (Read). A few photographs against none}
\casefacts{For each of eight breeds, say which output a classifier favors on photographs of that breed. During training some photographs were given the name of another breed.}
{Convolutional classifier of eight breeds, four look-alike cats and four look-alike dogs, with 327{,}112 parameters. Inside each group of four, names were moved either between two pairs of breeds or around a ring of four, and either most or only a few photographs were renamed. The task states this rule but not which case applies.}
{All weights, the rule above, and 8, 4, 2, 1 or 0 labeled photographs of each breed. Up to 50 runs, also on the supplied photographs.}
{About 50 photographs per breed.}
{The answer is right for every breed whose favored output stays the same when the hidden photographs are resampled.}
{Unseen data: real photographs.}
With no photographs 7 of 10 instances passed; with 1, 2, 4 and 8 photographs per breed 3, 4, 5 and 6 passed. Of these 50 instances, 42 are scored, 18 passes and 24 failures; seven passes with 2, 4 or 8 photographs are not scored, because a simple generic method solves those instances, and for one network no solution by permitted means is demonstrated. One photograph per breed turned six passes into misses and two misses into passes; on ten networks this difference is not statistically significant (exact test on the changed pairs, $p=0.29$), and we report it as an observation. With photographs \astra{} ran them through the classifier and averaged the answers per breed; according to our annotations it passed when it let the stated rule overrule some of these noisy measurements, and missed when it kept them. Without photographs it drew its own test pictures or read the weights. In the related family in which no version supplies photographs (seven tasks), 32 of 55 scored instances passed.

\casehead{Case F8 (Convert). Narrower first-layer filters: the same answers against a rising threshold}
\casefacts{Write a new first layer whose filters are 3 pixels high and 3 wide, in place of filters 3 high and 5 wide, so that the classifier names the same breed as before on photographs. A second family asks for filters of one row and five columns.}
{Convolutional classifier of 20 breeds, trained from scratch on about 150 photographs per breed. Only the first layer, 32 filters, may change.}
{All weights and a description of the training. No photographs. Up to 50 runs on at most 16 own images.}
{1{,}848 photographs of 37 breeds, 17 of which the classifier was never trained on.}
{Same breed as the original on at least 55\%, 65\%, 75\% or 78\% of the photographs (the required agreement of the version), with a 95\% margin.}
{Unseen data: real photographs.}
\astra{} used one recipe. It assumed a model of the statistics of natural photographs, replaced each filter by the smaller filter whose response is closest under that model, and rescaled it to the old strength. It almost never ran the result. On average its answers agreed with the original on 79 to 80\% of the photographs in every version, but single networks lie several points lower and the margin costs about two points more, so 10, 10, 8 and 5 of 10 instances passed (test for trend, $p=0.002$); 38 of these 40 instances are scored, 33 passes and 5 failures. Agreement on the 17 unseen breeds was 6.6 points lower on average than on the trained breeds. With one-row filters, 1 of 8 instances passed (six are scored, one pass and five failures), and the gap to the unseen breeds grew to 7.9 points; the passing program, run unchanged on the other instances, passed three more, and a procedure of 29 lines passes 6 of 8.

\clearpage
\section{Mathematical details and proofs}
\label{app:proofs}
This appendix states what each technique of Table~\ref{tab:tech} rests on and proves it; small worked examples close the parts where they help. Each part separates what is proved from what is assumed, because in several cases the mathematics is exact while its use on photographs is a modeling assumption. The formulas were extracted from the programs of the cited instances, derived again, checked numerically and reviewed independently. Throughout, $\relu(x)=\max(x,0)$, $\phi$ and $\Phi$ are the density and the distribution function of a standard normal variable, $\He_n$ are the probabilists' Hermite polynomials and $h_n=\He_n/\sqrt{n!}$.

\textbf{Propositions.} Proposition~1 (Case~S1, Appendix~\ref{app:cases}) is the content of Results~\ref{r:var} and~\ref{r:lmmse}, and Proposition~2 (Case~S2) follows from Results~\ref{r:herm} and~\ref{r:mehler}. The identities of Section~\ref{sec:mechanisms} are Result~\ref{r:fail}.

\subsection{A layer redesigned from stored statistics, with no data}
\emph{Task:} keep 12 to 28 of the 128 channels of the last convolution of a pet-photo classifier; no photographs.
\emph{What the network stores:} for each channel $i$, normalization constants such that the signal before the ReLU is
$\gamma_i(u_i+t_i)$ with $u_i$ of mean 0 and variance 1 on the training photographs, $t_i=\beta_i/\gamma_i$, $\gamma_i>0$.

\begin{wres}[Hermite series of a shifted ReLU]\label{r:herm}
For $U\sim N(0,1)$ and any $t$: $\relu(U+t)=\sum_{n\ge0}c_n(t)\,h_n(U)$ (the series converges in mean square) with
\[c_0=\phi(t)+t\Phi(t),\qquad c_1=\Phi(t),\qquad c_n=\frac{\phi(t)\,\He_{n-2}(-t)}{\sqrt{n!}}\quad(n\ge2).\]
\end{wres}
\emph{Proof.} $c_n=\E[f(U)h_n(U)]$ with $f(u)=\relu(u+t)$ and $h_n=\He_n/\sqrt{n!}$. Since $\He_n\phi=(-1)^n\phi^{(n)}$, integrating by parts
twice gives $\int f\,\He_n\,\phi=(-1)^{n-2}\int f''\phi^{(n-2)}$ for $n\ge2$, and $f''$ is a unit point mass at $-t$.
Hence $\int f\,\He_n\phi=(-1)^{n-2}\phi^{(n-2)}(-t)=\He_{n-2}(-t)\phi(-t)=\He_{n-2}(-t)\phi(t)$ by symmetry of $\phi$, and dividing by $\sqrt{n!}$ gives $c_n$. The cases $n=0,1$ are
$\E\relu(U+t)$ and $\Pr(U>-t)$. \qed

\begin{wres}[The covariance of two ReLU outputs needs only their correlation]\label{r:mehler}
If $(U,V)$ are jointly normal, each standard, with correlation $\rho$, then $\E[h_n(U)h_m(V)]=\rho^n$ if $n=m$ and $0$ otherwise.
Therefore
\[\Cov\big(\relu(U+s),\relu(V+t)\big)=\sum_{n\ge1}c_n(s)\,c_n(t)\,\rho^n ,\]
and for a whole layer with correlation matrix $R$ of the signals before the ReLU, the covariance of the outputs is
$K=\sum_{n\ge1}(c_nc_n^{\top})\circ R^{\circ n}$ (entrywise powers; $c_n$ is the vector with the entries $c_n(t_i)$).
\end{wres}
\emph{Proof.} $\E\big[e^{aU-a^2/2}e^{bV-b^2/2}\big]=e^{ab\rho}$. Expanding both sides in powers of $a$ and $b$, with
$e^{aU-a^2/2}=\sum_n a^n\He_n(U)/n!$, gives $\E[\He_n(U)\He_m(V)]=n!\,\rho^n\delta_{nm}$. Dividing by $\sqrt{n!\,m!}$ and inserting the series of
Result~\ref{r:herm} yields the formula. \qed\\
Results~\ref{r:herm} and~\ref{r:mehler} are known. The coefficients are those of a ReLU with bias~\citep{awasthi}, and the series is Theorem~1 of \citet{wright} for ReLU units. We include short proofs for completeness. If the physical signals are $\gamma_i(u_i+t_i)$ with gains $\gamma_i>0$, the covariance of channels $i$ and $j$ carries the factor $\gamma_i\gamma_j$. Cutting the series after $N$ terms costs at most $|\rho|^{N+1}\sqrt{T_N(s)\,T_N(t)}$, where $T_N(t)=\Var(\relu(U+t))-\sum_{n=1}^{N}c_n(t)^2$, because $T_N(t)=\sum_{n>N}c_n(t)^2$ and $|\rho|\le1$, by the Cauchy--Schwarz inequality applied to the unused coefficients.

\begin{wres}[The best next layer is a closed form]\label{r:ls}
Let $g$ be the 128 old pooled features and $g'$ the new ones, with model covariances $K=\Cov(g)$, $K_{ss}=\Cov(g')$,
$K_{xs}=\Cov(g,g')$, with $K_{ss}$ invertible, and let $J\succeq0$ weight the errors by what later layers read. Then
$\min_X\E\|J^{1/2}((g-\E g)-X(g'-\E g'))\|^2=\operatorname{tr}(JK)-\operatorname{tr}(JK_{xs}K_{ss}^{-1}K_{xs}^{\top})$,
attained at $X=K_{xs}K_{ss}^{-1}$.
\end{wres}
\emph{Proof.} Write $R(X)$ for the objective and $L=K_{xs}K_{ss}^{-1}$. Expanding the square gives $R(X)-R(L)=\|J^{1/2}(X-L)K_{ss}^{1/2}\|_F^2\ge0$, so no choice beats $L$; when $J$ is singular other choices tie with it. \qed

\emph{What \astra{} built from this.} New channels are linear mixes $T$ of the old signals with new thresholds $b$. Their
covariances follow from Result~\ref{r:mehler} (series cut at order 16 to 32), the next layer from
Result~\ref{r:ls} (with a small ridge), and $T,b$ are found by a quasi-Newton search on the remaining error. Pooling
and zero padding are built into $K$.

\emph{Proved:} the three results are exact. Given the Gaussian model and the new channels $(T,b)$, the next layer is the exact optimum of a least-squares problem (with a small ridge term). $T$ and $b$ themselves come from a numerical search, so they are a local optimum at best. No photo is used anywhere.\\
\emph{Assumed, not proved:} that the signals are jointly normal; that their correlations can be taken from the
similarity of the filters (which treats the layer below as uncorrelated); one fitted number for spatial correlation (0.493, fitted so that the predicted variances match the stored ones as closely as possible); and that the following ReLU can be ignored. \astra{}
hedged by averaging its objective over eight variants of the model.\\
\emph{Used in:} 42 of the 50 instances (all passed). Example: 16 channels kept, accuracy 0.92 points below the original,
lower end of the interval $-2.03$ against a limit of $-5$, no run.

\begin{wres}[The shift variant]\label{r:shift}
(i) Let $c>0$ (\astra{} used $c=32$ or $64$). If a compressed signal satisfies $u\ge-c$, then $O\,\relu(u+c)+b-c\,O\mathbf 1=Ou+b$; for any input the error is exactly $O\,\relu(-u-c)$. (ii) For $Z\sim N(\beta,\sigma^2)$ with $\sigma>0$, the best affine predictor of $\relu(Z)$ is
$\mu+p(Z-\beta)$ with $p=\Phi(\beta/\sigma)$ and $\mu=\beta p+\sigma\phi(\beta/\sigma)$. (iii) For a positive definite
weight $G$, the best rank-$12$ approximation of a matrix $M$ in the norm $\|G^{1/2}(\cdot)\|_F$ is
$G^{-1/2}(G^{1/2}M)_{12}$, the truncated singular-value decomposition taken after multiplying by $G^{1/2}$.
\end{wres}
\emph{Proof.} (i) $\relu(v)=v+\relu(-v)$. (ii) The slope is $\Cov(Z,\relu Z)/\sigma^2=\E[\mathbf 1\{Z>0\}]$ by Gaussian
integration by parts; the mean of $\relu Z$ is $\beta p+\sigma\phi(\beta/\sigma)$. (iii) Multiplication by $G^{1/2}$
keeps the rank, and the truncated SVD is optimal in the Frobenius norm (Eckart and Young). \qed\\
\emph{Used in:} 3 instances, all with 12 channels, all passed (on one more, both designs are mixed). \emph{Not proved:} that signals stay above $-c$ on real photographs; \astra{} wrote this caveat into its own design file.

\emph{In numbers.} If two pre-ReLU signals are jointly Gaussian with correlation $\rho$, the covariance after the ReLU is a known function of $\rho$. For two signals with mean 0 and variance 1 it is the series of Proposition~2, taken with both shifts $s$ and $t$ equal to zero, whose first terms are
\[
\Cov\big(\relu(U),\relu(V)\big)=0.2500\,\rho+0.0796\,\rho^{2}+0.0066\,\rho^{4}+0.0020\,\rho^{6}+\dots
\]
\emph{Worked example.} For unrelated signals ($\rho=0$) the covariance is 0. For copies ($\rho=1$) all terms of the series together add up to $0.341$, the variance $\tfrac12-\tfrac1{2\pi}$ of $\relu(U)$, as they must. For $\rho=0.5$ the first three terms give $0.1250+0.0199+0.0004=0.1453$, and the exact value, from the closed form $\big(\sqrt{1-\rho^2}+\rho\,(\pi-\arccos\rho)-1\big)/(2\pi)$~\citep{chosaul}, is also $0.1453$. For $\rho=0.8$ the value is $0.2544$.

\subsection{The unseen data rebuilt from numbers stored in the network}
\begin{wres}[Stored variances are linear equations in the covariance of the input]\label{r:var}
Let $x\in\R^d$ have mean $m$ and covariance $C$, and let unit $k$ compute $a_k^{\top}x+b_k$. Then its mean is
$a_k^{\top}m+b_k$ and its variance is $a_k^{\top}Ca_k=\sum_{i\le j}(2-\delta_{ij})a_{ki}a_{kj}C_{ij}$, which is
linear in the $d(d+1)/2$ entries of $C$. With $K$ units, every covariance $C$ is determined by its stored variances if and only if the $K\times d(d+1)/2$ coefficient matrix has full column rank. Without full rank a particular $C$ can still be determined when it lies on the boundary of the positive semidefinite cone, for example when two inputs are constant.
\end{wres}
\emph{Proof.} Linearity of the mean; $\Var(a^{\top}x)=a^{\top}Ca$. With full column rank the linear equations have at most one solution. Without it there is a symmetric matrix $\Delta\neq0$ with $a_k^{\top}\Delta a_k=0$ for all $k$, and for small $\varepsilon>0$ the covariances $I$ and $I+\varepsilon\Delta$ are both positive definite, differ and have the same stored variances. \qed\\
Recovering a covariance from quadratic measurements of this form is classical~\citep{chen}, and the fill-in of Result~\ref{r:lmmse} is the linear minimum mean squared error (Wiener) estimator~\citep{kay}.\\
In the task with deleted color values $d=12$ (a block of $2\times2$ pixels, three colors), so there are 78 unknowns and 96 equations,
and the measured rank was 78. No Gaussian assumption is needed here, only that the stored numbers are the true
averages over the training photographs.

\begin{wres}[The best linear fill-in for missing values]\label{r:lmmse}
Let only $y=Nx$ be delivered, with $NCN^{\top}$ invertible. Among all rules $\hat x=m+K(y-Nm)$, the mean squared error is smallest for
$L=CN^{\top}(NCN^{\top})^{-1}$, and any other $K$ is worse by exactly $\E\|(K-L)(y-Nm)\|^2$. Writing $AL$ into the
first layer and the constant $A(I-LN)m$ into its bias (or, with the opposite sign, into the stored mean) makes the network compute $A\hat x$.
\end{wres}
\emph{Proof.} The error $x-\hat x$ of $L$ is uncorrelated with $y$ because $CN^{\top}-LNCN^{\top}=0$; expand the
square. \qed\\
As a consequence, deleting a reading can never lower this best error, since a rule may always ignore a
reading. \emph{Not proved:} that a small error in the first layer keeps the breed decision.

\begin{wres}[What the first layer leaves open, the second layer fixes]\label{r:price}
For jointly normal $(U,V)$, each standard, with correlation $\rho$, $\Cov(\relu(U+a),\relu(V+b))=\int_0^{\rho}\Phi_2(a,b;r)\,dr$, where $\Phi_2(\cdot,\cdot;r)$ is the joint distribution
function at correlation $r$. Each stored second-layer variance is therefore a known smooth function of the unknown
correlations between first-layer units.
\end{wres}
\emph{Proof.} By Price's theorem~\citep{price} $\partial_\rho\E[f(U)g(V)]=\E[f'(U)g'(V)]$; here this is $\Pr(U>-a,V>-b)=\Phi_2(a,b;\rho)$, because $(-U,-V)$ has the same distribution as $(U,V)$,
and the covariance is $0$ at $\rho=0$. (The same function is the series of Result~\ref{r:mehler}; the two were checked
against each other.) \qed\\
\emph{Used in:} the hardest version of the input-distribution family (Case~S6; 5 of 5 instances), where 96 such equations fix 56 covariance
entries that the first layer cannot see, with a final error of $5\times10^{-14}$. \emph{Proved:} the equations.
\emph{Only numerical, and only local:} the Jacobian has full rank at the solution (smallest singular value 0.0037), so no other solution lies nearby; uniqueness over all covariances is not shown.

\begin{wres}[The mean color of the training photographs is in the biases]\label{r:bias}
Let a first-layer filter with weights $w_k$ be followed by normalization with stored mean
$\mu_k=\E[w_k^{\top}x]+b^0_k$, standard deviation $\sigma_k$, gain $\gamma_k$ and shift $\beta_k$. After folding, the file holds only the weights
$w'_k=(\gamma_k/\sigma_k)w_k$ and the bias $b'_k=\beta_k-w_k'^{\top}\E[x]$; the shifts $\beta_k$ themselves are no longer stored. If the mean image has one value $m_c$ per color, then \[b'_k+\sum_cS_{kc}\,m_c=\beta_k,\qquad S_{kc}=\textstyle\sum_{\text{taps }p}\pi_p\,w'_{kcp},\] where $\pi_p$ is the share of positions at which tap $p$ lies inside the image ($\pi_p=1$ without zero padding).
So the regression of $-b'$ on $S$ returns $m$ up to the scatter of the shifts $\beta_k$: if $\beta_k=\bar\beta+\varepsilon_k$,
and $[S,\mathbf 1]$ has full column rank, the least-squares estimate with an intercept errs by at most $\|[S,\mathbf 1]^{+}\|_2\,\|\varepsilon\|_2$. A fit without intercept returns $m-S^{+}\beta$ when $S$ has full column rank, where $S^{+}=(S^{\top}S)^{-1}S^{\top}$ is the least-squares inverse, so it is exact only if $S^{\top}\beta=0$; a small residual of the fit does not show this.
\end{wres}
\emph{Proof.} Folding $y=\gamma(w^{\top}x+b^0-\mu)/\sigma+\beta$ gives the first identity; a flat mean image gives the second;
the last is the standard perturbation bound for least squares. \qed\\
\emph{Proved:} the identity. \emph{Assumed:} that the shifts $\beta_k$ are nearly equal, or near zero when no intercept is fitted. On the examined instance \astra{} fitted without an intercept; its first fit left a residual of 0.062 against biases of size 0.263, so the assumption holds roughly, not exactly. Its final fit, with the padding shares $\pi_p$, gave the color $(0.475, 0.435, 0.389)$ on a 0 to 1 scale. On two other instances it fitted with an intercept.

\emph{Unproved smoothness assumption.} After folding, the response of filter $k$ on training photographs has variance
$w_k'^{\top}C\,w'_k=\gamma_k^2$ exactly. \astra{} assumed $\gamma_k\approx1$ for all 32 filters and fitted a
smoothness model of natural images (a power-law spectrum with 8 free numbers) to these 32 conditions. This is a sensible fit with more conditions than unknowns, but nothing guarantees the model family. In the task of Case~S16, among all 40 instances, \astra{} read photo statistics from the weights in either way on 20, of which 16 passed; the smoothness fit itself was used in 13 of them, of which 11 passed.

\emph{Worked example.} Let the inputs be the red and green values of a pixel, $x=(r,g)$. The unknowns are $c_{rr}=\Var(r)$, $c_{gg}=\Var(g)$ and $c_{rg}=\Cov(r,g)$. Suppose three units of the first layer have the weights $(1,0)$, $(0,1)$ and $(1,1)$, and the file stores the variances 4, 1 and 7 for them. Then
\[
c_{rr}=4,\qquad c_{gg}=1,\qquad c_{rr}+c_{gg}+2c_{rg}=7\ \Rightarrow\ c_{rg}=1 .
\]
Red and green are positively related, with correlation $1/\sqrt{4\cdot1}=0.5$.

\emph{Using it: filling in a missing value.} Suppose that only red is delivered. The best straight-line guess of green from red is
\[
\hat g=m_g+\frac{c_{rg}}{c_{rr}}\,(r-m_r)=m_g+0.25\,(r-m_r),
\]
where $m_r$ and $m_g$ are the means, which the file also determines. Without red, the error of guessing green has variance $c_{gg}=1$; with this guess it is $c_{gg}-c_{rg}^2/c_{rr}=0.75$. In the general formula of Proposition~1, $L^{\star}=CN^{\top}(NCN^{\top})^{-1}$, the observation matrix $N$ selects the values that are delivered. Here $N=(1\ \ 0)$, so $CN^{\top}=(4,1)^{\top}$, $NCN^{\top}=4$ and $L^{\star}=(1,\,0.25)^{\top}$. Red is kept, and green is filled in with the slope $0.25$.

\subsection{Arithmetic of residues in place of simulation}
\begin{wres}[Clock networks]\label{r:clock}
(i) \emph{Change of units.} If $h_{t+1}=\relu(Wh_t+b)$ and $D$ is diagonal and positive, then $z_t=D^{-1}h_t$ satisfies
$z_{t+1}=\relu(D^{-1}WDz_t+D^{-1}b)$.\\
(ii) \emph{A driven counter.} For an integer $m\ge1$, with $X'=\relu(X+U-mE)$, $E'=\relu(X+U-mE-(m-1))$, $U\in\{0,1\}$ and
$q=X-mE\in\{0,\dots,m-1\}$: $X'=q+U$, $E'=1$ exactly when $q+U=m$, and $X'-mE'=(q+U)\bmod m$. One carry per $m$ pulses.\\
(iii) \emph{Coincidences.} Events at $t\equiv a\pmod m$ and $t\equiv b\pmod n$ coincide if and only if
$\gcd(m,n)$ divides $a-b$, and then exactly once every $\operatorname{lcm}(m,n)$ steps. If it does not divide, the gate
that needs both never fires.\\
(iv) \emph{A clock from two units.} Let $P\ge2$ and $1\le f\le P-1$ be integers. With $r_0=P-1-f$, $p_0=0$, $r_{t+1}=\relu(r_t+1-Pp_t)$,
$p_{t+1}=\relu(r_t+3-P-Pp_t)$ one has $r_t=(P-1-f+t)\bmod P$ and $p_t=\mathbf 1\{r_t=P-1\}$, so $p$ pulses exactly at
$t=f+kP$.
\end{wres}
\emph{Proof.} (i) $\relu(du)=d\relu(u)$ for $d>0$. (ii) Direct substitution of the three cases $q+U<m-1$, $=m-1$, $=m$.
(iii) The Chinese remainder theorem. (iv) Induction over the three cases $r_t<P-2$, $r_t=P-2$, $(r_t,p_t)=(P-1,1)$. \qed\\
\emph{Proved:} everything; all quantities are integers, so the networks are exact at every step. Example: clocks with
periods 152 and 428 coincide every $16{,}264$ steps; four counters in a chain divide by 488, 234, 66 and 220, which
gives the carry periods $7{,}936{,}832$, $1{,}857{,}218{,}688$, $122{,}576{,}433{,}408$ and $26{,}966{,}815{,}349{,}760$ (the coincidence period $16{,}264$ times 488, and then times 234, 66 and 220).\\
\emph{Used in:} four task families; all 95 scored instances passed. The programs of 65 of them call routines for greatest common divisors, least common multiples or the remainder theorem.

\emph{A counter made of two ReLU units.} The following two units form a clock with period $P$ (the clock of Result~\ref{r:clock}, part~(iv); $P$ is a whole number of at least 2, and the clock starts with $p_0=0$ and $r_0$ between 0 and $P-2$):
\[
r_{t+1}=\relu(r_t+1-P\,p_t),\qquad p_{t+1}=\relu(r_t+3-P-P\,p_t).
\]
\emph{Worked example.} Take $P=3$ and start at $r_0=1$, $p_0=0$. Step by step:
\begin{center}\small
\begin{tabular}{@{}lrrrrrrrr@{}}\toprule
$t$ & 0 & 1 & 2 & 3 & 4 & 5 & 6 & 7\\\midrule
$r_t$ & 1 & 2 & 0 & 1 & 2 & 0 & 1 & 2\\
$p_t$ & 0 & 1 & 0 & 0 & 1 & 0 & 0 & 1\\\bottomrule
\end{tabular}
\end{center}
For instance, from $t=1$ to $t=2$: $r_2=\relu(2+1-3\cdot1)=0$ and $p_2=\relu(2+3-3-3)=\relu(-1)=0$. The unit $r$ counts $0,1,2,0,1,2,\dots$ and the unit $p$ gives a pulse at $t=1,4,7,\dots$, that is, whenever $t\equiv1\pmod3$. All values are whole numbers, so nothing is ever rounded. This works because as long as $p_t=0$, the first unit adds 1 at every step. The second unit is positive only when $r_t=P-2$, so $p$ becomes 1 at the step at which $r$ reaches $P-1$. One step later the term $-P\,p_t$ sets $r$ back to 0, and $p$ returns to 0.

\emph{A second example.} A clock pulses at $t\equiv1\pmod4$ and another at $t\equiv3\pmod6$. Here $\gcd(4,6)=2$ divides $1-3=-2$, so they meet, every 12 steps: at $t=9,21,33,\dots$ (check: $9=2\cdot4+1=1\cdot6+3$). Now change the second clock to $t\equiv2\pmod6$. Then $\gcd=2$ does not divide $1-2=-1$, and the two never meet: the first clock pulses only at odd times and the second only at even times. A unit that waits for both will never fire. This is a proof, valid for all time; running the network can only show that the unit has not fired so far.

\subsection{Counting inside units that saturate}
Because $\tanh$ is nonlinear and saturates, a count stored in it drifts, and finite-precision GRUs cannot count without bound~\citep{weiss}. The construction below is Richardson extrapolation~\citep{richardson} applied to the scale of the unit. \astra{} stored the same count $u$ at several small scales
$s_0>s_1>\dots>s_{k-1}$ and read it back as $F(u)=\sum_i\alpha_i\tanh(s_iu)/s_i$.

\begin{wres}[Canceling nonlinearity across scales]\label{r:lanes}
Let the scales be distinct, $q_i=(s_i/s_0)^2$ and $\alpha_i=\prod_{j\ne i}q_j/(q_j-q_i)$. Then $\sum_i\alpha_i=1$, $\sum_i\alpha_iq_i^p=0$ for
$1\le p\le k-1$, and $\sum_i\alpha_iq_i^k=(-1)^{k-1}\prod_jq_j$. Consequently, with $\tanh x=\sum_m\tau_mx^{2m+1}$,
\[F(u)-u=(-1)^{k-1}\tau_k\Big(\prod_is_i^2\Big)u^{2k+1}+O(u^{2k+3}),\]
as $u\to0$; that is, all error terms up to order $2k-1$ cancel.
\end{wres}
\emph{Proof.} The $\alpha_i$ are the values at $0$ of the Lagrange basis polynomials for the nodes $q_i$, so
$\sum_i\alpha_ip(q_i)=p(0)$ for every polynomial of degree below $k$. For $p(x)=x^k$ the interpolation error formula
gives $\sum_i\alpha_iq_i^k=-\prod_j(-q_j)=(-1)^{k-1}\prod_jq_j$. Now
$F(u)=\sum_m\tau_mu^{2m+1}s_0^{2m}\sum_i\alpha_iq_i^m$. \qed

\begin{wres}[Two scales, with a bound valid for all $u$]\label{r:two}
For $F_s(u)=\big(8\tanh(su)-\tanh(2su)\big)/(6s)$: $0\le F_s'\le1$ and $|F_s(u)-u|\le\frac{8}{15}s^4|u|^5$.
\end{wres}
\emph{Proof.} With $v=\tanh(su)$, $1-F_s'(u)=\frac43\,v^4(2+v^2)/(1+v^2)^2$. This is at least $0$ and at most $\frac83(su)^4$; it also increases with $|v|$ and equals $1$ at $|v|=1$, so $F_s'\ge0$. Integrate from $0$. \qed

\begin{wres}[Why the count stays exact for a million steps]\label{r:induct}
Suppose the stored count starts at the true integer $n_0$, each step computes $c_{t+1}=F(c_t+\text{pulses}_t)$, and on an interval containing both $c_t+\text{pulses}_t$ and the true integer $n_t+\text{pulses}_t$ at every step, $F$ does not expand distances and
$|F(u)-u|\le\delta$. Then, as long as the pulses are the true ones, after $T$ steps the stored count differs from the true integer by at most
$T\delta$. If $T\delta<\frac12$, every threshold test at a half-integer gives the right answer, so the pulses and the
logic are exact.
\end{wres}
\emph{Proof.} Induction: the error at step $t+1$ is at most the error at step $t$ (non-expansion) plus $\delta$, and
while it is below $\frac12$ the pulses are the true ones. \qed\\
\begin{wres}[An exact formula and a bound for the error]\label{r:exact}
Write $\bar q_i=s_i^2$ and $t_j=4/(\pi^2(2j+1)^2)$. For every real $u$,
\[u-F(u)=u\sum_{j\ge0}2t_j\prod_{i}\frac{t_ju^2\bar q_i}{1+t_ju^2\bar q_i}.\]
Hence $u-F(u)$ is odd and increasing in $u$, $F'\le1$ everywhere, and for $|u|\le C$
\[|u-F(u)|\le\delta=C^{2k+1}\Big(\prod_i\bar q_i\Big)\big[\tau_k^{+}-2t_0^{k+1}(1-D^{-1})\big],\qquad D=\prod_i(1+t_0C^2\bar q_i),\]
where $\tau_k^{+}=|\tau_k|=\sum_j2t_j^{k+1}$ is the absolute value of the coefficient of $x^{2k+1}$ in $\tanh x$.
\end{wres}
\emph{Proof.} The partial fractions of $\tanh$ give $\tanh z/z=\sum_j2t_j/(1+t_jz^2)$, and $\sum_j2t_j=1$. The weights of Result~\ref{r:lanes} satisfy $1-\sum_i\alpha_i/(1+x\bar q_i)=\prod_i x\bar q_i/(1+x\bar q_i)$: after multiplication by $\prod_i(1+x\bar q_i)$ this reads $\sum_i\alpha_i\prod_{l\ne i}(1+x\bar q_l)=\prod_i(1+x\bar q_i)-x^k\prod_i\bar q_i$, and both sides are polynomials of degree below $k$ that agree at the $k$ points $x=-1/\bar q_i$. Put $x=t_ju^2$ and sum over $j$. Every summand is non-negative and increasing in $|u|$, which gives the monotonicity and $F'\le1$. For the bound, keep the denominator $D$ in the summand $j=0$ and bound every other denominator below by one. \qed\\
With $k=6$, scales $2^{-8},\dots,2^{-13}$ and $C=139.01$ this gives $\delta<2.625\times10^{-13}$ and $2^{20}\delta<2.753\times10^{-7}$; on this range $F'>0$ as well, so $F$ is non-expanding and Result~\ref{r:induct} applies.

\emph{Numbers.} Two scales, $s=1/6144$, 256 steps, counts up to 139: $T\delta\le0.005$ by Result~\ref{r:two}. Six scales $2^{-8},\dots,2^{-13}$,
$2^{20}$ steps, counts up to 139: the leading term of Result~\ref{r:lanes} is $3.06\times10^{-13}$ at $u=139.01$, the value \astra{}'s program uses; the true maximum of $u-F(u)$ on the range, computed by us with 80-digit arithmetic, is $2.62\times10^{-13}$, $F$ is non-expanding there, and $T\delta=2.75\times10^{-7}$. In the real network every step also rounds. \astra{}'s program bounds this rounding ($1.0\times10^{-11}$ per step, about 40 times the error above) and arrives at $1.1\times10^{-5}$ for the whole run, still far below $\frac12$.\\
\emph{Proved:} Results~\ref{r:lanes} to \ref{r:exact}, for exact arithmetic. \astra{}'s own program bounds the error by the leading term only and multiplies by the number of steps; Result~\ref{r:exact} supplies the missing bound on the remainder and the non-expansion.\\
\emph{Used in:} 3 of the 40 instances of this task. The other 37 also passed with other designs, including 17 exact digital
counters built from gates, 11 with one unit kept in its linear range and a separate pulse unit, 8 with a unit that
resets itself, 1 with a geometric counter.

\emph{Two scales.} Store the count at two scales, $s$ and $2s$, and combine the readings as
\[
F(u)=\frac{8\tanh(su)-\tanh(2su)}{6s}.
\]
The combination is chosen so that the terms with $u^3$ cancel. The value $\tanh(su)$ falls short of $su$ by about $(su)^3/3$, and $\tanh(2su)$ falls short of $2su$ by eight times as much. Eight times the first value minus the second therefore leaves no such term, and dividing what is left, about $6su$, by $6s$ gives back $u$. The error that remains is much smaller, about $\tfrac{8}{15}s^4u^5$ in size.

\emph{Worked example.} With $s=0.01$ and $u=20$: $\tanh(0.2)=0.197375$ and $\tanh(0.4)=0.379949$, so $F=(8\cdot0.197375-0.379949)/0.06=19.9842$. The error fell from $0.26$ to $0.016$ at the same scale. This is the idea of Richardson extrapolation. We combine two imperfect answers whose errors have a known form so that the leading errors cancel. With six scales instead of two, five kinds of error cancel, and the count stays exact for a million steps.

\subsection{One memory shared by two recurrent networks}
\begin{wres}[]\label{r:gru}
(i) Let $x_{t+1}=Fx_t+Gu_t+d$, $x_0=0$. Every $x_t$ lies in the span $\mathcal V$ of the columns of
$L,FL,F^2L,\dots$ with $L=[G\ d]$. If $U$ has orthonormal columns spanning $\mathcal V$, then $q_t=U^{\top}x_t$ obeys
$q_{t+1}=U^{\top}FUq_t+U^{\top}Gu_t+U^{\top}d$ and $x_t=Uq_t$ exactly.\\
(ii) A recurrent unit of the GRU type with both gates held at $\frac12$ and state $h=\delta q$ computes
$h'=h/2+\tanh(\delta w)/2$ with $w=(2A-I)q+2(Bu+d)$. This equals $\delta(Aq+Bu+d)$ up to an error of at most
$\delta^3\|w\|_\infty^3/6$ per coordinate.
\end{wres}
\emph{Proof.} (i) $\mathcal V$ contains $L$ and is mapped into itself by $F$. (ii) $|\tanh v-v|\le|v|^3/3$. \qed\\
Part (i) is the reachable-subspace argument of linear systems theory~\citep{kalman}. Two networks that read the same inputs are one stacked system, so (i) applies to both at once, provided the units in question are driven only by each other and by the inputs (signals from other units count as further inputs $u_t$). The 96 nearly linear
units of the two networks span only 16 directions (the residual of the invariance test in one examined instance was
$3.6\times10^{-14}$).\\
\emph{Proved:} the reduction and the encoding error of one update. The full nonlinear network over many steps is validated by the hidden test and not by a proof. \emph{Approximate:} the original units apply $\tanh$ to their
state (error at most $|v|^3/3$), and at the tightest width 16 directions are cut to 12.\\
\emph{Used in:} all 70 instances, which all passed, down to 44 units.

\subsection{Unusual units made to behave like ordinary ones}
\begin{wres}[Tiny signals turn units with slope one at zero into wires]\label{r:wire}
For every real $x$ and $\varepsilon>0$:\quad
$\big|\frac{1}{\varepsilon}\big(\varepsilon x+\sin^2(\varepsilon x)\big)-x\big|\le\varepsilon x^2$;\quad
$\big|\frac{1}{\varepsilon}\frac{\varepsilon x}{1+|\varepsilon x|}-x\big|\le\varepsilon x^2$;\quad
$\big|\frac{1}{\varepsilon}\sin(\varepsilon x)-x\big|\le\varepsilon^2|x|^3/6$.
For the chain sine, soft-sign $\sigma(z)=z/(1+|z|)$, sine and $|q|\le M$: $|\sin(\sigma(\sin(\varepsilon q)))/\varepsilon-q|\le\varepsilon
M^2+\varepsilon^2M^3/3$.
\end{wres}
Scaling an activation near a point where its derivative is non-zero to obtain the identity is a standard device of approximation theory~\citep{kidger}.\\
\emph{Proof.} $\sin^2z\le z^2$; $z-z/(1+|z|)=z|z|/(1+|z|)$; $|\sin z-z|\le|z|^3/6$; add the three errors of the chain, in which every signal is at most $\varepsilon M$ in size because $|\sin z|\le|z|$ and $|\sigma(z)|\le|z|$.
\qed\\
With $\varepsilon=2^{-44}$ and $M=100$ the chain error is $5.7\times10^{-10}$ against a tolerance of $10^{-5}$.
This is the bound for one carried signal; \astra{}'s bounds for the three outputs were $6\times10^{-10}$ to $2.3\times10^{-9}$. Scaling by a power of two is exact in floating point. \emph{Not covered by \astra{}'s bounds:} rounding inside the units. The graded errors of the two examined instances ($3.5\times10^{-12}$ and $2.8\times10^{-11}$) show it is harmless.

\begin{wres}[Absolute values carry a linear function]\label{r:abs}
$\relu(z)=\frac12(z+|z|)$. If $V^{\top}h+d\ge1$ for all $h$ in a box, then $C\,|V^{\top}h+d|-Cd=CV^{\top}h$ on that box.
The smallest value of a coordinate of $V^{\top}h$ over the box $[\mathrm{lo},\mathrm{hi}]$ is $V_+^{\top}\mathrm{lo}+V_-^{\top}\mathrm{hi}$ with $V_+=\max(V,0)$ and $V_-=\min(V,0)$, so $d$ can be computed exactly.
A linear map of rank $r$ thus needs only $r$ absolute-value units.
\end{wres}
\emph{Proof.} Immediate. \qed\\
\emph{Proved:} this carrier. \emph{Not proved, only measured on the supplied examples:} which units may be treated
as always on or off and which near-copies may be merged in the classifier-head task. That part passed all 52 scored instances on the
hidden data; one further instance, for which no solution by permitted means is demonstrated, was missed.

\begin{wres}[A ReLU from snake units]\label{r:cos}
On $[-R,R]$: $\relu(z)=\frac z2+\frac R4-\frac{2R}{\pi^2}\sum_{k\ \mathrm{odd}}\frac{\cos(k\pi z/R)}{k^2}$. A snake unit
$\psi(z)=z+\sin^2z$ satisfies $c\,\psi(az)=caz+\frac c2-\frac c2\cos(2az)$. Sixteen such units with frequencies
$a_k=\pi k/(2R)$, weights $C_k=4R/(\pi^2k^2)$ for the odd $k\le29$, and the first term split into the pair $\pm a_1$
with weights $(C_1\pm\delta)/2$, $\delta=(\frac12-\sum_{k\ge3}C_ka_k)/a_1$, plus the constant $\frac R4-\frac12\sum_kC_k$ (all sums over the odd $k\le29$), give $\relu(z)+e(z)$ with $|e(z)|\le\frac{2R}{\pi^2}\sum_{k\ge31,\ \mathrm{odd}}k^{-2}\approx0.0034\,R$ on $[-R,R]$.
\end{wres}
\emph{Proof.} The Fourier series of $|z|$ on $[-R,R]$; $\sin^2=\frac12(1-\cos2\,\cdot)$; the pair $\pm a_1$ adds the slope
$\delta a_1$ without changing the cosines, and $\delta$ is chosen so that all slopes sum to $\frac12$. \qed\\
\emph{Not proved:} that signals stay inside $[-R,R]$. \astra{} took $R$ as 1.5 times the largest value seen on 16 supplied example images (1.2 or 1.25 times in other instances). All 10 instances passed.

\begin{wres}[The output layer as the last ReLU]\label{r:mmm}
For three scores $(B+\alpha t,\ B,\ -2B-\alpha t)$ with $\alpha>0$, ``maximum minus mean'' equals $B+\alpha\relu(t)$
whenever $3B+\alpha t\ge0$.
\end{wres}
\emph{Proof.} The scores sum to zero, so the mean is $0$; the third score is at most $B$ exactly when $3B+\alpha
t\ge0$. \qed\\
\astra{} proved the condition over the whole input box with interval bounds (margins 19, 17 and 44 in one examined instance).

\subsection{Overwritten units rebuilt from the network's symmetry}
\begin{wres}\label{r:twin}
Write each healthy unit $i$ (with $a_i\ne 0$) of a one-hidden-layer ReLU network as $a_i\relu(w_i^{\top}x+b_i)=s_i\relu(r_i^{\top}x+c_i)$ with
$s_i=\operatorname{sign}(a_i)$, $r_i=|a_i|w_i$, $c_i=|a_i|b_i$. Suppose the healthy set of triples $(s_i,r_i,c_i)$ is
closed under a group of input permutations acting by $\pi\cdot(s_i,r_i,c_i)=(s_i,\pi r_i,c_i)$. Suppose also that the damage left the signs of all outgoing weights intact. (i) A unit whose orbit is larger than the number of units of its sign cannot be healthy. (ii) Suppose in addition that every unit not excluded by (i) is healthy and that the outgoing weights of the damaged units are intact and non-zero. If the orbits of the remaining units contain as many distinct triples as there are units, they are the whole healthy set, and writing each missing triple into a damaged slot of the same sign restores the output
function for every input.
\end{wres}
\emph{Proof.} (i) A healthy unit brings its whole orbit, all of the same sign. (ii) Every healthy completion contains
these orbits, and there is no room for more; the output is the sum of the triples' contributions, which does not depend
on the order of the slots. \qed\\
The extra assumption holds in the examined instance. The task specifies exactly eight damaged units there, and (i) excludes eight. That the units of an invariant shallow ReLU network are organized by its symmetry group is known~\citep{agrawal}. \emph{Used in:} all 40 annotated instances (31 of them scored); three further scored instances also passed. In one examined instance 8 of 36 units were damaged, the group had 24 elements and the
output error was zero on all 2,048 graded inputs.

\subsection{Units that can never switch off}
\begin{wres}\label{r:on}
If the features satisfy $g\ge0$ and a unit has $w\ge0$ and $b\ge0$, then $\relu(w^{\top}g+b)=w^{\top}g+b$. Let $M$
be the matrix by which all such units of a head together act on its $K$ outputs, that is, their outgoing weights times their incoming weights; their biases only add a constant to each output. With the column shift $s_k=\max(0,-\min_rM_{rk})$, the rows $M_r+s$
and $s$ are non-negative, so $\relu((M_r+s)g)-\relu(sg)=M_rg$ for all $g\ge0$: $K$ output rows cost $K+1$ ReLU units.
If only the top class is graded, subtracting the same number from all scores of a head changes nothing, and the extra
unit can be dropped.
\end{wres}
\emph{Proof.} All arguments of the ReLUs are non-negative. \qed\\
Removing or merging units whose ReLU is provably always active or always inactive is the basis of lossless compression~\citep{serra}.\\
\emph{Proved:} exactness in real arithmetic for every non-negative feature vector, hence on all photographs (the grader computes in single precision, where the scores differ by about $3\times10^{-5}$). The non-negativity of the
features follows from the architecture and was not stated in most versions of the task. In the tightest version this
gives $18+17+8+8+1=52$ units.\\
\emph{Used in:} all 70 scored instances. \emph{The same step without the proof} is the main failure in the
photo-specialist task (Section~\ref{sec:mechanisms}). There the belief that the units are always active was a guess, and the exact cost of a wrong guess
is $-o_j\relu(-z_j)$ per unit (Result~\ref{r:fail}(ii)).

\subsection{Hidden integer rows}
\begin{wres}\label{r:rat}
(i) $\relu(tz)=t\relu(z)$ for $t>0$, so a unit with incoming row $tq$ can be replaced by the row $q$ with its outgoing
weights multiplied by $t$, and units with the same $q$ can be merged by adding their outgoing weights. (ii) If a real number $y$ is closer than $1/(2D^2)$ to a fraction $p/q$ with $q\le D$, that fraction is the only one with denominator at most $D$ that close; it is recovered by choosing the nearest fraction with denominator at most $D$, which the continued-fraction algorithm finds.
\end{wres}
\emph{Proof.} (i) Positive homogeneity. (ii) Two different fractions with denominators at most $D$ differ by at least
$1/D^2$, giving uniqueness. Moreover, since $|y-p/q|<1/(2D^2)\le1/(2q^2)$, Legendre's theorem guarantees that $p/q$ appears as a convergent of the continued-fraction expansion of $y$. \qed\\
\emph{What makes the answer rigorous} is not (ii) but the check that follows: \astra{} bounded the difference between its
integer network and the original over the whole input box in exact fractions, including an allowance for
floating-point rounding ($4.3\times10^{-10}$ against a limit of $10^{-6}$).\\
\emph{Used in:} the 15 instances of three versions. On the 5 instances of the fourth version (integers up to 7) \astra{} searched over integer rescalings instead. All 20 passed.

\subsection{One correction that is best on average over an unknown change}
\begin{wres}\label{r:post}
Let the new sensor deliver $x'=Ax+\beta$ where $(A,\beta)$ is unknown with some probability law (here: the posterior
given a few readings), and let $x$ be independent of $(A,\beta)$ with mean $\mu$ and covariance $C$. Among all rules $z\approx P'(x'-\mu')$ for the
normalized input $z=P(x-\mu)$, the mean squared error is smallest for \[\mu'=\E[A\mu+\beta],\qquad P'=P\,C\,\E[A]^{\top}\big(\E[ACA^{\top}]+\Cov(A\mu+\beta)\big)^{-1},\] provided the matrix in parentheses is invertible.
\end{wres}
\emph{Proof.} This is the best linear predictor of $z$ from $x'$: $\Cov(z,x')=PC\,\E[A]^{\top}$ and, by the law of total
covariance, $\Cov(x')=\E[ACA^{\top}]+\Cov(A\mu+\beta)$. \qed\\
\emph{Why the same formula passes with 16 readings and fails with 4.} The correction contains $\E[A]$. When the
readings leave the rotation angle of a pair of sensor channels open, with both signs about equally likely, the block of $\E[A]$ for that pair shrinks toward a multiple of the identity and the correction no longer undoes its rotation. In the checked failing
instance the sign of one angle had probability 0.5005. The rule is optimal for squared error and still useless for a
classifier that needs a sharp inverse.\\
\emph{Used in:} all 20 instances with 64 or 16 readings, which all passed, and most instances with 8 or 4 readings, where 5 of 20 passed; the 15 misses are not scored, because no solution by permitted means is demonstrated for those instances. On four further instances with 4 readings (Case~S12), 2 of 4 passed.

\subsection{Proofs in place of tests}
\begin{wres}[Rounding in a recurrent state machine]\label{r:round}
The construction used has $2n$ state units for a machine with $n$ states: in exact arithmetic unit $u_{j,c}$ is 1 when the machine has just read the bit $c$ and reached state $j$, and 0 otherwise. With $T_c$ the 0/1 transition table for the bit $c$ (one 1 in each row) and $x$ the next bit, held in an input register, one step computes
\[u'_{j,0}=\relu\Big(\sum_i(T_0)_{ij}\,(u_{i,0}+u_{i,1})-x\Big),\qquad u'_{j,1}=\relu\Big(\sum_i(T_1)_{ij}\,(u_{i,0}+u_{i,1})+x-1\Big).\]
Let $e_t$ bound the total deviation $\sum_{j,c}|u_{j,c}-u^{\star}_{j,c}|$ of the state units from their exact values $u^{\star}_{j,c}$, $\rho$ the deviation of an input
register, $n$ the number of states and $\gamma$ the rounding factor of one inner product: a computed pre-activation $a^{\top}h+b$ is off by at most $\gamma\,(|a|^{\top}|h|+|b|)$. Then
$e_{t+1}\le2e_t+2n\rho+\gamma\,(2(1+e_t)+2n(1+\rho)+n)$, because each state unit feeds weights of total size 2 and ReLU
does not expand distances. Iterating over all steps bounds the error of the state for every input at once. The output is one further inner product of the state, so its error is at most the largest read-out weight times the final $e_t$, plus one rounding term of the same form; the bound quoted below includes this step.
\end{wres}
\emph{Proof.} Each row of $T_0$ and of $T_1$ holds one 1, so each of the two sums, added over $j$, counts every state unit once. Before rounding, the $2n$ pre-activations therefore deviate by at most $2e_t$ in total through the state units and by $2n\rho$ through the register. The rounding of a pre-activation is at most $\gamma$ times the sum of the absolute values of its terms; added over the $2n$ units, these sums are at most $2(1+e_t)$ for the state units, whose exact values add up to 1, $2n(1+\rho)$ for the register (since the exact bit satisfies $x^\star\in\{0,1\}$, so $|x|\le1+\rho$) and $n$ for the biases $-1$. ReLU does not expand distances. \qed\\
In one examined instance (14 states, 16 bits, 15 steps) this gives $3.55\times10^{-8}$ against a limit of $10^{-7}$, in
exact fractions, with no network run; the measured error over all 65,536 inputs was $1.8\times10^{-11}$.
\emph{Proved by the program:} the rounding bound. \emph{Justified only in code comments (not formally verified):} that the exact-arithmetic network implements the state machine.

\begin{wres}[A certificate for the largest output]\label{r:cert}
Suppose that on the relevant region the output is $y(x)=\kappa+c^{\top}x+\mathrm{cap}-\sum_n\pi_n|A_nx+b_n|$ with constants $\kappa$ and $\mathrm{cap}$ and with
$\pi_n>0$. If a point $p$ of that region solves $Ap=-b$ and there are multipliers with $A^{\top}\lambda=c$ and $|\lambda_n|\le\pi_n$, then $y(x)\le\kappa+\mathrm{cap}-\lambda^{\top}b$ for all $x$ in that region, with equality at $p$.
\end{wres}
\emph{Proof.} $c^{\top}x=\sum_n\lambda_n(A_nx+b_n)-\lambda^{\top}b\le\sum_n\pi_n|A_nx+b_n|-\lambda^{\top}b$, with equality at $p$, where every $A_np+b_n$ is zero. \qed\\
\astra{} reached this form by merging rows that are copies or mirror images ($\relu(-z)=\relu(z)-z$; 384 units became 140
nodes) and checked $p$, $\lambda$ and both conditions in exact fractions, together with two facts the certificate needs: $p$ lies inside the input box, and outside the region the output stays below (at most 5.93 against 11.375). The certified maximum $372743/32768$ equals
the graded value exactly. In deeper versions parts of the bound rest on floating-point solvers.

\begin{wres}[A unit that creates a kink cannot be removed]\label{r:kink}
In an open convex region where all first-layer units keep their state, the pre-activations of the second layer are affine,
$z=R[x;1]$. If row $j$ changes sign inside the region, has a non-zero output weight, and no other row has the same
zero set, the output has a jump in slope across $\{z_j=0\}$. If removing a set $S$ of first-layer units changes $R$
so that no row has this zero set any more, the reduced network is smooth there, so it differs from the original, and
$S$ is not removable.
\end{wres}
\emph{Proof.} Removing $S$ leaves the pre-activations of the other first-layer units as they were, so these units keep their state in the region and the second-layer pre-activations of the reduced network are again affine there. Constant rows bend nothing. The zero sets of the other rows, before and after the removal, are finitely many hyperplanes different from $\{z_j=0\}$, so some point $x_0$ of $\{z_j=0\}$ in the region lies on none of them. Near $x_0$ the reduced network is affine, while the gradient of the original jumps by the output weight of row $j$ times $\nabla z_j\neq0$. A function with a slope jump across a hyperplane is not equal to one that is smooth across it. \qed\\
Dual certificates and proof-producing verification of ReLU networks are established~\citep{wongkolter,isac}, and kinks are used to identify units when networks are reverse-engineered~\citep{rolnick}.\\
Together with linear programs that prove which second-layer units are never active, this pins down the full answer.
All 9 scored instances matched. In the examined instance the regions come from floating-point linear programs, while the rank tests and the bounds for the never-active units are exact.

Equality of two networks was proved by the same kind of collapse as in Result~\ref{r:cert}. Every unit is reduced to a canonical exact expression, and equal expressions are merged. Where the two networks differ, \astra{} exhibited an input on which they differ. All 15 scored instances passed.

\subsection{Small exact circuits}
\begin{wres}\label{r:gad}
(i) \emph{One-point detector.} For bit strings $x,v\in\{0,1\}^d$,
$\relu\big((2v-1)^{\top}x+1-\|v\|_1\big)=\relu(1-\text{number of positions where }x\ne v)$, which is 1 if $x=v$ and 0
otherwise. With a mask it detects a pattern on the masked bits.\\
(ii) \emph{First match wins.} If $m_1,\dots,m_R\in\{0,1\}$, then $s_i=\relu(m_i-\sum_{k<i}m_k)$ is 1 exactly for the
first $i$ with $m_i=1$.\\
(iii) \emph{An exact bit from an integer score.} For integers $s,\theta$:
$\relu(s-\theta+1)-\relu(s-\theta)=\mathbf 1\{s\ge\theta\}$.\\
(iv) \emph{Parity.} For $N\in\{0,\dots,m\}$: $\relu(N)-2\relu(N-1)+2\relu(N-2)-\dots\pm2\relu(N-m+1)=N\bmod 2$.\\
(v) \emph{Switching off.} If $0\le h\le U$ and $\pi\in\{0,1\}$, then $\relu(h-(U+1)\pi)$ equals $h$ for $\pi=0$ and $0$
for $\pi=1$.\\
(vi) \emph{Repurposing a unit via a rank-one update.} Replacing row $j$ of a layer by
$\gamma q^{\top}$ with zero bias is the rank-one update $e_j[\gamma q^{\top}-A_j\mid-b_j]$, and it changes the output
scores by exactly $B_{:,j}\big(\relu(\gamma q^{\top}g)-h_j(g)\big)$.
\end{wres}
\emph{Proof.} (i) $(2v-1)^{\top}x=\|v\|_1-\#\{x\ne v\}$ on bit strings. (ii) to (v) by cases. (iv): the left side has slope
$+1$ on $[0,1]$, then $-1$, $+1$, \dots, so it alternates between 0 and 1 at integers. (vi) Only row $j$ changes.
\qed\\
\emph{Used in:} the rule-list tasks (all 15 scored instances passed with zero error), the switch-off task (all 25 scored instances passed; zero error on all
65,536 inputs), and the one-breed edit (14 of the 21 instances on which it was used passed, scored or not). In (vi) the mathematics says exactly what changes; whether the new
detector fires on the right photographs is measured on four examples, not proved.

\subsection{Reading the sign of an edit: exact margins, but no guarantee}
\begin{wres}\label{r:sign}
If an edit $D$ is added to the output layer, the score difference between the target breed $c$ and a rival $r$ changes
by exactly $(D_c-D_r)^{\top}h$ at features $h\ge0$. If the row $F_c$ has a positive entry, the direction $u=\relu(F_c)/\|\relu(F_c)\|_2$ maximizes the target score over non-negative vectors of Euclidean length 1 (Cauchy--Schwarz, since $F_c^{\top}u\le\relu(F_c)^{\top}u$ for $u\ge0$). But no rule that reads only the weights can decide in general whether recognition
goes up. For the two-class network $z_c=h_1$, $z_r=h_2+1$ and the edit $(1,-2)$ on the target row, a photo with
$h=(0.75,0)$ flips from misclassified to correctly classified, while one with $h=(3,1.8)$ flips from correctly classified to misclassified.
\end{wres}
\astra{}'s rule (read the edit on the breed's strongest weights) was right on all 300 edits of the 30 scored instances. The
counterexample shows that this is a property of these networks and edits, not a theorem.

\subsection{Three identities behind the failures}
\begin{wres}[Three identities]\label{r:fail}
(i) For $h\ge0$ and rows $v_a,v_b$, with $\max(v_a,v_b)$ taken element-wise, $\max(v_a,v_b)^{\top}h\ge\max(v_a^{\top}h,v_b^{\top}h)\ge\frac12(v_a+v_b)^{\top}h$, and $\max(v_a^{\top}h,v_b^{\top}h)-\frac12(v_a+v_b)^{\top}h=\frac12|v_a^{\top}h-v_b^{\top}h|$; the same holds with bias terms by appending $1$ to $h\ge0$. (ii) $o_jz_j-o_j\relu(z_j)=-o_j\relu(-z_j)$. (iii) If $\|\tilde z(x)-z(x)\|_\infty\le\epsilon$ and the margin $m(x)=z_c(x)-\max_{j\ne c}z_j(x)$ of the winning coordinate $c=\arg\max_j z_j(x)$ satisfies $m(x)>2\epsilon$, then $\tilde z$ and $z$ have the same largest coordinate. Consequently $\Pr[\arg\max\tilde z\ne\arg\max z]\le\Pr[m\le2\epsilon]+\Pr[\|\tilde z-z\|_\infty>\epsilon]$ for inputs from any distribution.
\end{wres}
\emph{Proof.} (i) The element-wise maximum dominates both rows and $h\ge0$, which gives the first inequality; the rest is $\max(p,q)=\frac12(p+q)+\frac12|p-q|$. (ii) $z-\relu(z)=-\relu(-z)$. (iii) Let $c$ be the winner under $z$. For every $j\ne c$, $\tilde z_c-\tilde z_j\ge z_c-z_j-2\epsilon\ge m(x)-2\epsilon>0$. The probability statement is the union bound over the two ways in which the premise can fail. \qed\\
\emph{Use.} Part~(i) explains why an averaged row fails on look-alike classes. It is weakest where one member of the pair is recognized most clearly (Case~F1, Figure~\ref{fig:mechanism}a). Part~(ii) is the exact cost of folding a ReLU on a wrong guess that it never clips (Case~F2, Figure~\ref{fig:mechanism}b). Part~(iii) is the standard margin argument~\citep{neyshabur}; it shows what a check in weight space leaves open. It bounds neither the share of inputs with a small margin nor the share on which the score error is large. An exact rewrite has $\epsilon=0$ on every input.

\emph{Worked example.} Let the first breed be recognized by dark ears and the second by a striped coat, with the rows $v_a=(6,0)$ and $v_b=(0,6)$ on the features (ears, stripes). A photograph with dark ears and no stripes has $h=(1,0)$. The old scores are 6 and 0.
\begin{itemize}
\item The \emph{average} row is $(3,3)$ and scores $3$. A rival class that scores 4 now wins, and the photograph is lost.
\item The \emph{element-wise maximum} is $(6,6)$ and scores $6$, as high as the stronger old class.
\end{itemize}
In general the average falls short of the stronger class by exactly half the difference of the two old scores, here $\tfrac12|6-0|=3$. The loss is largest exactly where one breed is recognized most clearly. The maximum never falls short, because increasing a weight cannot lower a score when the features are not negative.

\subsection{Artifacts left by file generation}
\label{app:leakmath}
The two results below explain routes that the solver found in the task files and not in the task; the instances on which they were used are not scored (Section~\ref{sec:exploits}). In a verification task the proposed repairs were stored at a finer number format than the hidden healthy network, and in 61 of 110 instances \astra{} used this trace to read the healthy network back from the proposals (Result~\ref{r:f32}); in one instance it also divided the weight decay out of all 392 training steps. Related exploits are known~\citep{depue,shao,alug}, and seeds of pseudo-random generators have been recovered in other settings~\citep{machtle}; solving for the state of the generator~\citep{argyros} from the weights of a network has, to our knowledge, not been reported.

\begin{wres}[The random seed can be read back from weights that training left unchanged]\label{r:seed}
Suppose a weight was drawn as $w=b(2u-1)$ with $u=(z\bmod 2^{24})/2^{24}$, where $z$ is a 32-bit output of the
Mersenne Twister generator and $b=1/\sqrt{\text{fan-in}}$.\\
(i) A weight that training never changed lies on the grid $-b+k\cdot2b\,2^{-24}$, $k$ an integer, and reveals the low 24 bits of $z$. (As a stored single-precision number it is exactly on the grid when $b$ is a power of two; otherwise $k$ is confirmed by regenerating the weight.)\\
(ii) The generator's update and output steps are linear over bits. Each such weight therefore gives 24 linear
equations over the two-element field in the 19,968 bits of the generator's state.\\
(iii) The seeding rule $s_i=1812433253\,(s_{i-1}\oplus(s_{i-1}\gg30))+i \pmod{2^{32}}$ can be run backward. The solved state is the one after the generator's first update; running that update backward returns the seeded words except part of the first, and the seed follows from the second word by one backward seeding step.
\end{wres}
\emph{Proof.} (i) By the formula for $w$. (ii) Shifts, masks and exclusive-or are linear over bits. (iii) The
multiplier is odd, hence invertible modulo $2^{32}$; and $v=s\oplus(s\gg30)$ is undone by $s=v\oplus(v\gg30)$, because
shifting twice by 30 leaves nothing of a 32-bit word. \qed\\
State recovery of this generator from truncated outputs is classical~\citep{argyros}. In one instance the equations of 833 untouched values, together with the 31 linear relations that hold in every state produced by an update, gave a system of rank 19,968, equal to the number of unknowns, so the state and then the seed were unique. The 31 relations come from the last step of the update, which is carried out in place. With $s$ the state before and $s'$ the state after the update, this step sets $s'_{623}=s'_{396}\oplus\operatorname{mix}\big((s_{623}\,\&\,\mathtt{0x80000000})\mid(s'_0\,\&\,\mathtt{0x7fffffff})\big)$, where $\operatorname{mix}(y)=(y\gg1)\oplus(y\bmod2)\cdot\mathtt{0x9908b0df}$. The map $\operatorname{mix}$ is linear over bits and invertible, because the top bit of its constant is 1, so each of the low 31 bits of $s'_0$ is a fixed linear function of the bits of $s'_{623}\oplus s'_{396}$. These relations are needed. The low 31 bits of the first state word enter no output but the first, so the 24-bit observations alone cannot reach this rank. On other instances \astra{} scanned seeds directly. On two of them it executed hand-assembled machine code in memory at several million seeds per second, locating the seed after 0.3 and 1.2 billion trials. Three ordinary scans in another task covered all $2^{32}$ seeds and found nothing.

\begin{wres}[A hidden original network can be read back from rounding]\label{r:f32}
Let $H$ and $B$ be matrices stored in single precision and let a candidate be published as $B+E$ with $E=s(H-B)$ in double
precision. Then $X(s')=B+E/s'$ equals $H$ up to double-precision rounding, and so lies on the single-precision grid up to that rounding, when $s'=s$. At $s'=s/k$ for a small integer $k$, $X=kH-(k-1)B$ is an integer combination of grid numbers and part of its entries are on the grid again.
\end{wres}
\emph{Proof.} At $s'=s$, $X=B+(H-B)=H$ up to double-precision rounding; in general $X=H+(s/s'-1)(H-B)$. \qed\\ \emph{Observed, not proved:} at strengths unrelated to $s$ the entries fall evenly between grid points, at an average distance of a quarter of the grid spacing. The true strength is therefore the lowest point of the scan, not its only dip.\\
In one examined instance the average distance was $1.7\times10^{-7}$ grid steps at $s=0.925$, between $0.12$ and $0.18$ at simple fractions of it, and about $0.25$ elsewhere, over 1,548 entries. On a second instance \astra{} recovered the scale differently. Weights that receive no gradient are only
multiplied by $1-10^{-3}\cdot5\cdot10^{-4}$ at each training step, and \astra{} divided them back through all 392 steps
in single precision, checking each step. The result matched one candidate's edit with scale $1.9125$ and a relative residual of $2\times10^{-11}$ ($3\times10^{-12}$ in a second, independent comparison).

\end{document}